\documentclass[pdflatex,sn-mathphys-num]{sn-jnl}% Math and Physical Sciences Numbered Reference Style
\usepackage{graphicx}%
\usepackage{multirow}%
\usepackage{amsmath,amssymb,amsfonts}%
\usepackage{amsthm}%
\usepackage{mathrsfs}%
\usepackage[title]{appendix}%
\usepackage{xcolor}%
\usepackage{textcomp}%
\usepackage{manyfoot}%
\usepackage{booktabs}%
\usepackage{algorithm}%
\usepackage{algorithmicx}%
\usepackage{algpseudocode}%
\usepackage{listings}%

\usepackage{lineno}
\usepackage{hyperref}
\usepackage{fancyhdr}

\newcommand{\R}{\mathbb{R}}

\newcommand{\vardim}{d}

\newcommand{\tensor}[1]{\boldsymbol{#1}}

\newcommand{\gray}[1]{\textcolor{gray}{#1}}

\usepackage[normalem]{ulem}
\newcommand{\mathsout}[1]{%
  \mathchoice
    {\text{\sout{$\displaystyle #1$}}}%
    {\text{\sout{$\textstyle #1$}}}%
    {\text{\sout{$\scriptstyle #1$}}}%
    {\text{\sout{$\scriptscriptstyle #1$}}}%
}
\DeclareRobustCommand{\removed}[1]{%
  \ifmmode
    {\mathsout{#1}}%
  \else
    {\sout{#1}}%
  \fi
}

\theoremstyle{thmstyleone}%
\theoremstyle{thmstyletwo}%

\theoremstyle{thmstylethree}%

\begin{document}
%\linenumbers
\title[eFDPC with shuffler anonymization]{Evolving Federated Density Peak Clustering with Shuffler-Based Anonymization}

\author*[1]{\fnm{Miha} \sur{Ožbot}}\email{miha.ozbot@fe.uni-lj.si}

\author[2]{\fnm{Seiichi} \sur{Ozawa}}\email{ozawasei@kobe-u.ac.jp}

\author[1]{\fnm{Igor} \sur{Škrjanc}}\email{igor.skrjanc@fe.uni-lj.si}

\affil*[1]{%
  \orgdiv{Faculty of Electrical Engineering},
  \orgname{University of Ljubljana},
  \orgaddress{\street{Tržaška cesta 25}, \city{Ljubljana},
              \postcode{1000}, \country{Slovenia}}%
}

\affil[2]{%
  \orgdiv{Center for Mathematical and Data Sciences},
  \orgname{Kobe University},
  \orgaddress{\street{1-1 Rokkodai-cho, Nada-ku}, \city{Kobe},
              \postcode{657-8501}, \country{Japan}}%
}

\abstract{
In this paper, we present a novel one-shot federated clustering approach based on evolving fuzzy clustering and shuffler-based anonymization. In federated clustering, each round reveals additional model information to the server and can increase the opportunity to infer properties of local datasets, even when raw samples are never shared. Consequently, fewer client--server exchanges and unlinkability are important, so transmitted information cannot be attributed to any single client. Data-stream clustering algorithms, which learn in a single pass over the data, align well with the one-shot federated learning paradigm, where a single communication round can be sufficient. The proposed Evolving Federated Density Peak Clustering (eFDPC) method combines evolving fuzzy data-stream clustering and density peak clustering in a federated setting. Each client independently evolves micro-clusters from private data mini-batches through cluster addition and merging. To limit communication, density peak clustering is applied to the learned micro-clusters and only the selected peaks are transmitted, rather than the full local models. A shuffler-based layer is introduced between the clients and the server to anonymize contributions prior to reconstruction and global peak selection. We consider a peak-only variant, \mbox{eFDPC-$\mathcal{P}$}, as well as \mbox{eFDPC-$\mathcal{G}$} with server-side label propagation. Experiments on synthetic and real-world benchmark datasets show that eFDPC outperforms recent federated clustering baselines in one-shot settings and scales to higher-dimensional, non-IID data.
}


\maketitle

\fancyhf{}                       % clear header and footer
\renewcommand{\headrulewidth}{0pt}
\fancyfoot[C]{\parbox{\textwidth}{\centering\footnotesize
  \copyright\ 2026 The Author(s). This article is published open access under a
  Creative Commons Attribution 4.0 licence (CC BY 4.0).\\
  Miha O\v{z}bot, Seiichi Ozawa, Igor \v{S}krjanc, ``Evolving Federated Density
  Peak Clustering with Shuffler-Based Anonymization,'' \emph{Evolving Systems},
  vol. 17, art. 123, 2026. DOI:
  \href{https://doi.org/10.1007/s12530-026-09884-0}{10.1007/s12530-026-09884-0}}}
\thispagestyle{fancy}
\clearpage

\section{Introduction}\label{sec:introduction}

\par A fundamental challenge in recent years is the sharing of information between different entities in sectors like manufacturing, government, finance, and healthcare, \emph{etc.}, while maintaining data privacy~\cite{McMahan_Moore_Ramage_Hampson_Arcas_2017}. Federated Learning (FL) addresses the challenge of learning from distributed data that cannot be shared directly due to concerns such as privacy, economics, and security, \emph{inter alia}. The contributors of the data, typically referred to as clients, benefit from shared information. However, supervised learning poses a challenge in a federated setting, as most real-world data are unlabeled. Additionally, the interpretability of the federated model is crucial in fields like finance and healthcare, where understanding the underlying system and trusting the model are essential~\cite{Hu_Jiang_Dong_Liu_He_2026, Ngo_Fang_Rumbut_Wang_2024}. Federated clustering (FC) offers a solution to this problem by enabling clustering analysis and the sharing of expert knowledge without requiring access to the data itself. One client might possess a large amount of data, while another might have expert knowledge. Unsupervised Federated Learning can enable the utilization of previously unavailable data to increase information utilization. The data are commonly non-independently and non-identically distributed (non-IID) among clients~\cite{Li_Sahu_Zaheer_Sanjabi_Talwalkar_Smith_2020, Li_Huang_Yang_Wang_Zhang_2019}, \emph{i.e.}, the task is shared, but local distributions differ, with each client having a distinct number of samples and class representation. None of the clients possess the global cluster centroids, so they must collaborate by sharing their local models to obtain a consistent global solution. Collaboration in creating the federated model is beneficial to each individual client, from one that desires information but does not have data, to a client that possesses the data but lacks the resources or knowledge to utilize it, to the client that has biased or partial data.

\par Traditional clustering methods require specifying the number of clusters, typically focusing on minimizing the average distance between samples and cluster centers or maximizing the likelihood of the observed data under the model. However, this can be problematic in a federated learning setting, where the number of clusters may not be known in advance. The number of clusters may vary for each client, and the clusters and their number may change over time, \emph{e.g.}, fraudulent financial transactions~\cite{Kanamori_Abe_2022}, malicious network activity~\cite{Skrjanc_Ozawa_Ban_Dovzan_2018}, \emph{etc.} Clients would benefit from shared information to address new challenges quickly. We propose that evolving data stream clustering, which was designed to adapt clusters incrementally over time and add new ones when necessary, is well suited to address these challenges.

\par Evolving Fuzzy Systems (EFS)~\cite{Skrjanc_Iglesias_Sanchis_Leite_Lughofer_Gomide_2019} are designed to learn from streaming data, where each sample arrives sequentially and is subject to concept drift. As data characteristics change over time, these algorithms adapt clusters incrementally and introduce new clusters as needed. The system aims to extract maximum information from each individual sample before discarding it. While a limitation of this approach is that processing samples sequentially does not scale with available processing capacity and results in significant memory access overhead, we modify the usual streaming clustering approach to learn from mini-batches of data instead, which aligns better with federated learning. Evolving systems typically generate a large number of micro-clusters that capture local structure in the data, so a subsequent selection step is required to identify representative peaks and remove redundant or outlier micro-clusters. While the evolving-systems literature provides mature mechanisms for adding, merging, splitting, removing, and forgetting clusters, using full multivariate covariance matrices becomes costly as dimensionality increases. Gaussian micro-clusters using diagonal covariances scale linearly with data dimensionality, and the final clustering can be done by selecting and grouping these micro-clusters.
\par Density Peak Clustering (DPC)~\cite{Rodriguez_Laio_2014} identifies cluster centers as points that are both locally dense and well separated from higher-density points. Classical DPC relies on global pairwise comparisons to rank densities and to determine nearest-higher relationships, which is incompatible with federated settings where raw samples cannot be centralized. Each client first performs evolving clustering on private mini-batches that results in local micro-clusters. DPC is then applied locally to the learned micro-clusters to select a small set of representative peak candidates, rather than sharing all micro-clusters or raw samples.

\par Although raw samples are never shared, transmitted peak candidates may still be attributable to individual clients across rounds, potentially leaking information about local data distributions. To prevent attribution, we introduce a shuffler-based anonymization layer between clients and the server. Each client splits each peak statistic into additive shares, shufflers permute the received share-messages, and the server reconstructs peaks only after mixing. The server thus obtains an unlinkable pool of candidate peaks, performs a second DPC step to select the global peak set, and can optionally propagate peak labels to obtain a full grouping over the reconstructed representatives.

\begin{figure}[!t]
    \centering
    \includegraphics[width=1\linewidth]{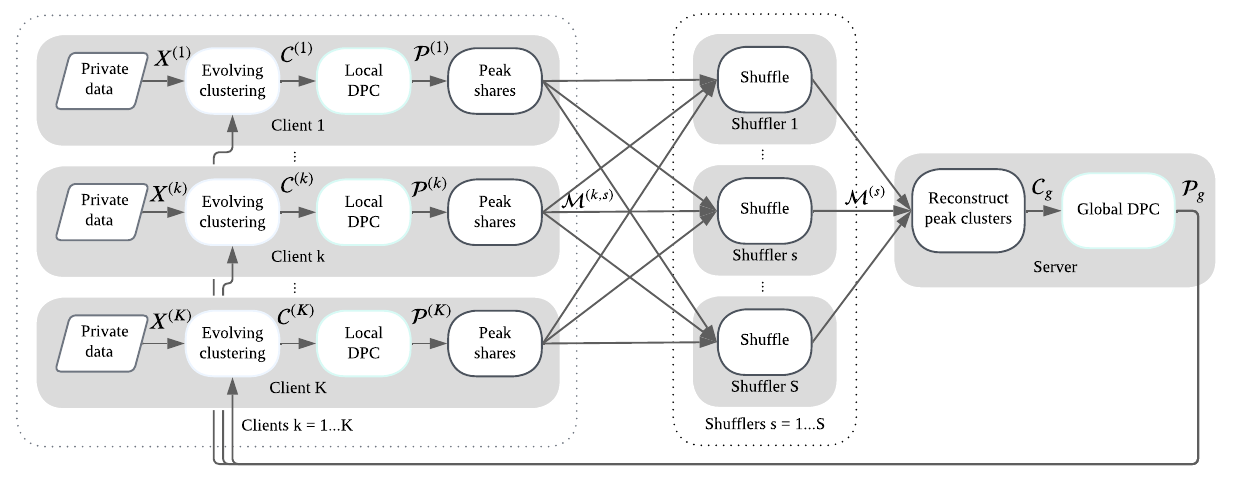}
    \caption{Overview of the proposed eFDPC pipeline. (1) Each client learns evolving micro-clusters from its private data using mini-batch updates. (2) Each client applies a local DPC step on its micro-clusters to select a small set of local peaks. (3) The selected peak statistics are split into additive shares and sent to multiple shufflers. Each shuffler permutes the received messages and forwards them to the server. (4) The server reconstructs the peak statistics by matching share identifiers and summing the shares, obtaining an unlinkable pool of candidate peaks. (5) The server performs a second DPC step on the reconstructed candidates to select the global peak set, and (6) broadcasts the global peaks back to clients for the next communication round.}
    \label{fig:federated_clustering}
\end{figure}

\par The proposed Evolving Federated Density Peak Clustering (eFDPC) algorithm {is a one-shot federated adaptive clustering framework for heterogeneous decentralized data and} is illustrated in Fig.~\ref{fig:federated_clustering}. The main contributions of the paper are:
\begin{itemize}
\item An improved evolving fuzzy clustering methodology that operates on mini-batches instead of the traditional online-learning setting, together with a novel measure for computing cluster overlap for merging.
\item A novel density peak clustering approach that operates on micro-clusters rather than data samples.
\item A federated clustering approach that combines evolving clustering and density peak clustering with a shuffler layer to provide client anonymity.
\end{itemize}

\par The paper is structured as follows. Section~\ref{sec:introduction} introduces the problem and reviews related work on federated clustering, evolving systems, and Density Peak Clustering. Section~\ref{sec:materials_and_methods} describes the proposed evolving systems methodology, the DPC peak selection, and the federated clustering framework. The experimental setup and results are presented in Section~\ref{sec:experimentation}. A discussion is provided in Section~\ref{sec:discussion}, and the conclusions are presented in Section~\ref{sec:conclusion}.
\subsection{Related Literature}
\label{sec:related_literature}
\par Prototype-based federated clustering methods like $k$-means~\cite{Dennis_Li_Smith_2021, Corcuera_Barcena_Marcelloni_Renda_Bechini_Ducange_2024, Pan_Sima_Prakash_Rana_Milenkovic_2022, Xu_Chen_Chao_Zhang_2024} and fuzzy $C$-means (FCM)~\cite{Corcuera_Barcena_Marcelloni_Renda_Bechini_Ducange_2024, Stallmann_Wilbik_2022, Hu_Qin_Shen_Pedrycz_Liu_Liu_2024} were proposed by researchers to address different challenges of federated learning, \emph{i.e.}, one-shot learning~\cite{Dennis_Li_Smith_2021, Xu_Chen_Chao_Zhang_2024}, non-IID data~\cite{Dennis_Li_Smith_2021, Stallmann_Wilbik_2022}, multiview data~\cite{Hu_Qin_Shen_Pedrycz_Liu_Liu_2024}, unlearning client data~\cite{Pan_Sima_Prakash_Rana_Milenkovic_2022},
vertical feature partitioning~\cite{Corcuera_Barcena_Marcelloni_Renda_Bechini_Ducange_2024}. 
A key methodological distinction of these clustering approaches lies in the aggregation process at the server. For instance, clusters may have one-to-one relationships, the clustering process may be repeated at the server using the cluster centers as samples, or clusters may be iteratively merged based on specific selection criteria. One of the first proposed federated clustering method was ${k}$-FED~\cite{Dennis_Li_Smith_2021}, which modified the classic $k$-means method for federated learning.
In ${k}$-FED, local clusters are generated using a $k$-means method and then aggregated at the server by iteratively selecting the most separated cluster centers from the local clusters, followed by performing $k$-means on all local centers. This approach allows each local client to have a different number of clusters, though the number must be specified in advance. Similarly, a one-shot learning Federated Centroid Aggregation (FeCA) method was proposed in~\cite{Xu_Chen_Chao_Zhang_2024}, also based on $k$-means, which combines clusters by iteratively selecting the cluster with the largest radius and merging it with nearby clusters. In contrast, the $k$-means++ clustering approach was used in MUFC~\cite{Pan_Sima_Prakash_Rana_Milenkovic_2022} to improve cluster initialization for the local clients and to aggregate the local clusters at the server. The LossLess Federation $k$-means (LLF-CM) and FCM (LLF-FCM) methods proposed in~\cite{Corcuera_Barcena_Marcelloni_Renda_Bechini_Ducange_2024} address horizontal and vertical data partitioning. The initial set of cluster centers is selected from one of the clients. The server then aggregates the local counts of points and the sum of sample values for each cluster, updating the global cluster centers accordingly. In FedFCM~\cite{Stallmann_Wilbik_2022}, clusters are also initialized by one of the clients. Each client clusters its data using FCM with the global clusters as initial values and communicates the updated cluster centers and sum of memberships or support. Notably, the authors highlight the challenge of selecting the appropriate number of clusters as an unsolved problem. A federated multiview fuzzy $C$-means consensus prototypes clustering (FedMVFPC) approach was proposed in~\cite{Hu_Qin_Shen_Pedrycz_Liu_Liu_2024}, demonstrating that achieving consensus among clusters from multiple views is beneficial. In federated multiple imputation-based fuzzy (FeMIFuzzy) method~\cite{Ngo_Fang_Rumbut_Wang_2024} clients vote on the optimal number of clusters using the Xie-Beni index. Those federated prototype-based clustering methods are based on iterative procedures, repeating local reclustering and server aggregation over multiple communication rounds. Even when a fixed message size enables secure aggregation, repeated rounds increase communication cost and can make prototype trajectories linkable across rounds. 

\par Fuzzy systems were designed to handle data uncertainty and noise, making them suitable for federated learning scenarios where data can be non-IID across devices~\cite{Zhang_Shi_Chang_Lin_2023}. Fuzzy models have recently been proposed for distributed and federated learning. In this context, clustering approaches are used to define the antecedent structure of the model. Fuzzy neural networks have been applied in distributed learning~\cite{Shi_Lin_Chang_Ding_Shi_Yao_2021}, using $k$-means clustering, and in distributed semi-supervised learning~\cite{Shi_Zhang_Cao_Tanveer_Lin_2022}, employing the distributed FCM clustering method. Furthermore, fuzzy models commonly rely on a consequent local linear model for regression and classification tasks. A federated learning system (Fed-FRBS) based on the Takagi-Sugeno Fuzzy Rule-Based System was proposed in~\cite{Corcuera_Barcena_Ducange_Ercolani_marcelloni_Renda_2022}, with an emphasis on Explainable AI. Fuzzy c-means clustering is used to determine the model antecedents and weigh contributions to the federated learning model based on prediction quality. FedFNN~\cite{Zhang_Shi_Chang_Lin_2023} proposed a federated fuzzy neural network that learns antecedent Gaussian centers and variances through back-propagation. To handle non-IID data and uncertainties, an evolutionary algorithm is used. EFS extend the fuzzy modeling approach by integrating unsupervised clustering techniques to incrementally learn and partition data in an online manner~\cite{Skrjanc_Iglesias_Sanchis_Leite_Lughofer_Gomide_2019}.

\par EFS incrementally cluster data to create membership functions for fuzzy rules and can add, merge, split, and remove clusters based on different criteria. These methods are designed for online classification~\cite{Angelov_Xiaowei_Zhou_2008,Pratama_pedrycz_Lughofer_2018}, regression~\cite{Lughofer_Skrjanc_2023}, time series forecasting~\cite{Wang_Liu_Chen_2022, Dovzan_Logar_Skrjanc_2015, Lughofer_Pratama_2022}, and clustering~\cite{skrjanc_2020, Blazic_Skrjanc_2020, Strzinar_skrjanc_Pregelj_2025} in evolving environments or streaming data. Several evolving clustering approaches were proposed based on Gaussian distribution~\cite{Dovzan_Logar_Skrjanc_2015, skrjanc_2020, Wang_Liu_Chen_2022, Blazic_Skrjanc_2020}, Cauchy distribution~\cite{Skrjanc_Ozawa_Ban_Dovzan_2018}, data clouds ~\cite{Angelov_Gu_Principe_2018, Bezerra_Costa_Guedes_Angelov_2020}
data granules~\cite{Garcia_Leite_Skrjanc_2020}, interval prototypes~\cite{Strzinar_skrjanc_Pregelj_2025}, among others. Each cluster definition also requires a tailored methodology for incremental learning, as well as a cluster merging metric and computation. Specifically, evolving Gaussian clustering is used to determine information granulation in~\cite{Wang_Liu_Chen_2022}. The granules are compared based on the Kullback–Leibler (KL) divergence, which is a measure of distribution dissimilarity. An evolving Cauchy clustering (eCauchy) method was proposed in~\cite{Skrjanc_Ozawa_Ban_Dovzan_2018} for monitoring and identification in cybersecurity intrusion detection. The evolving volume-based Gaussian Clustering (eGauss+) approach~\cite{skrjanc_2020} was designed to learn from data streams for clustering and regression tasks. An incremental version of Gustafson–Kessel clustering method was proposed in~\cite{Dovzan_Logar_Skrjanc_2015} to identify the antecedent clusters of an evolving fuzzy system for modeling a waste-water treatment facility. Incrementally learned granules of data were used to describe the antecedent of an evolving granular fuzzy-rule-based modeling method in~\cite{Garcia_Leite_Skrjanc_2020}. An autonomous learning multi-model (ALMMo) evolving classifier, which uses the data cloud-based Cauchy clusters model, was proposed in~\cite{Angelov_Gu_Principe_2018}. Similarly, AutoCloud ~\cite{Bezerra_Costa_Guedes_Angelov_2020} relies on Typicality and Eccentricity of the samples instead of defining a prototype shape. An approach using the Recursive least squares was used in~\cite{Blazic_Skrjanc_2020} to cluster streaming data based on the error of the model prediction. Evolving learning from data chunks instead of per sample was explored in~\cite{Lughofer_Pratama_2022} similar to mini-batches with a changing number of samples. Recently, an evolving federated learning method was introduced in~\cite{Ozbot_Ozawa_Skrjanc_2024} for supervised learning, that uses evolving fuzzy systems that can dynamically add clusters, starting from scratch. Most evolving clustering methods are designed for centralized streams and rely on sample-level access and second-order statistics, with merge and evaluation steps that are sensitive to tuning. Extending these mechanisms to federated settings is challenging due to limited communication, heterogeneous local evolution, and privacy constraints on transmissible parameters.

\par DPC has recently received increased attention due to its ability to identify density peaks and model manifold-like cluster structures. The core idea of DPC is to compute a local density estimate for each sample and the distance from that sample to its nearest neighbor of higher density. Samples that have simultaneously high density and large distance to higher-density samples are selected as cluster centers, and remaining samples are assigned by following the density gradient \cite{Tobin_Zhang_2024}. Recently, the DPC research focuses more on local density for neighboring points rather than relying on global distances, especially in high-dimensional settings. A common strategy is to restrict computations to $k$-nearest neighbors (kNN), which reduces the impact of irrelevant distances in high dimensions and makes distance estimates more local and comparable \cite{Yang_Xiao_2024, Xie_Zhong_Tian_Hu_Pan_2026, Du_Ding_Jia_2016}. DPC and its variants have been applied to intrusion detection systems (IDS) as well \cite{Wang_Zhang_Shen_Wang_Deng_Zhao_2025}. Further extensions combine kNN with dimensionality reduction (e.g., principal component analysis {(}PCA{)}) to stabilize local neighborhoods and improve center selection \cite{Du_Ding_Jia_2016}. Robust variants of DPC have also been proposed to improve stability and reduce sensitivity to density perturbations \cite{Ding_Li_Ding_Xu_Guo_Wu_2025}. To reduce computational cost, one approach first approximates the dataset using representative granular-ball centers and then performs clustering on these representatives instead of on raw samples \cite{Ding_Li_Ding_Xu_Guo_Wu_2025}. In federated or distributed settings, several designs combine local summarization with server-side peak selection. For example, one-shot pipelines can perform client-side $k$-means summarization and then apply SNN-DPC center selection at the server \cite{Wang_Pang_Wang_Pedrycz_2025}. Horizontal federated DPC (HFDPC) has also been studied as a direct adaptation of DPC to horizontally partitioned data \cite{Ding_Li_Xu_Guo_Ding_Wu_2025}. To mitigate the so-called domino effect in assignments caused by unstable nearest-higher relations, some methods explicitly restrict the distance-to-higher computation to the $k$ nearest neighbors \cite{Chen_Du_Zhou_Wu_Wang_2022}. While this improves stability, it typically introduces an additional parameter that constrains the acceptable neighbor range (or neighborhood validity), which must be selected carefully for each dataset \cite{Chen_Du_Zhou_Wu_Wang_2022}.

\section{Methods}
\label{sec:materials_and_methods}
\par In this section, we present the methodological components of the eFDPC algorithm, including evolving clustering, density peak clustering, and federated learning.
\subsection{Evolving Clustering}\label{subsec:evolving_methodology}
\par Let $\tensor{X} = [\tensor{x}_i]^\top \in \mathbb{R}^{N \times \vardim}$ denote the data matrix, where $\mathcal{I}=\{1,\dots,N\}$ is the index set of data points, $N\in\mathbb{Z}^+$ is the number of samples, and $\vardim\in\mathbb{Z}^+$ is the feature dimension. The objective of the local clustering is to partition $\tensor{X}$ into a set of micro-clusters
\begin{equation}\label{eq_local_clusters}
\mathcal{C}=\{\mathcal{C}_1,\dots,\mathcal{C}_C\} =\big\{(\tensor{\Sigma}_c,\tensor{\mu}_c,n_c)\,\big|\,c\in\{1,\dots,C\}\big\},
\end{equation}
where each micro-cluster is defined by a diagonal covariance matrix $\tensor{\Sigma}_c\in\mathbb{R}^{\vardim\times \vardim}$, a center $\tensor{\mu}_c\in\mathbb{R}^{\vardim}$, and the {support} $n_c\in\mathbb{R}_{+}$. Evolving clustering incrementally adapts the number of micro-clusters $C$ to represent all samples while suppressing redundant overlap by merging. This can be summarized by the representation condition:
\begin{equation}
d_{c^* i}^2 \le \kappa_\sigma^2, \quad \forall i\in\mathcal{I},\quad \exists\, c^*\in\{1,\dots,C\},
\end{equation}
where $d_{c^*i}^2$ is the {squared} Mahalanobis distance from sample 
$\tensor{x}_i$ to its nearest cluster center, and 
$\kappa_\sigma \in \mathbb{R}^+$ is the representation threshold.
The representation condition is complemented by the post-merge 
separability condition:
\begin{equation}\label{eq_separability_postmerge}
\kappa_{pq} > \kappa_m, \qquad \forall\,p\neq q, \quad p,q\in\{1,\dots,C\},
\end{equation}
where $\kappa_{pq}\in(1,\infty)$ denotes the overlap score between 
micro-clusters $\mathcal{C}_p$ and $\mathcal{C}_q$, and 
$\kappa_m\in(1,\infty)$ is the merging threshold.
\par Evolving clustering consists of the following main steps:
\textit{membership computation}, \textit{incremental clustering}, \textit{cluster adding mechanism}, and \textit{cluster merging mechanism}~\cite{Skrjanc_Iglesias_Sanchis_Leite_Lughofer_Gomide_2019}. The evolving per-sample learning methodology is not directly applicable to federated learning since updates cannot be sent for each sample due to privacy and communication cost.  eFDPC evolving operations are performed in parallel for all samples in the mini-batch and clusters simultaneously.
\subsubsection{Membership computation} The Mahalanobis distance is used to calculate the distance between each sample $\tensor{x}_i$ and each cluster $\mathcal{C}_c$:
\begin{align}
d^2_{ci} & = (\tensor{x}_{i}{-}\tensor{\mu}_c)^{\top}\tensor{\Sigma}_c^{-1}(\tensor{x}_{i}{-}\tensor{\mu}_c),\label{eq_mahalanobis_distance} \\
{u}_{ci} & = \mathrm{exp}\left({-}{{d}_{ci}^2}/{\vardim}\right),\label{eq_membership}
\end{align}

where $d^2_{ci}\in\R^+$ is the {squared} Mahalanobis distance, $\tensor{u}_{c} = [u_{ci}] \in \R^{N}$ is the degree of association of the samples $\tensor{X}$ with the cluster $\mathcal{C}_c$. The scaling factor $1 / \vardim$ is used to improve high-dimensional stability $\mathbb{E}[d_{ci}^2] = \vardim$~\cite{Etherington_2019, Vaswani_Shazeer_Parmar_Uszkoreit_Jones_Gomez_Kaiser_Polosukhin_2017}.

\subsubsection{Cluster adaptation}

\par Let $\mathcal{I}_{c^*}$ denote the index set of newly assigned samples that satisfy the representation constraint:
\begin{equation}\label{eq_incremented_samples}
\mathcal{I}_{c^*} = \left\{i \;\middle|\; c^* = \arg\min_{c} d_{ci},\; u_{c^* i} > \exp\!\left(-\kappa_\sigma^2/\vardim\right) \right\}.
\end{equation}
The parameters of the nearest assigned cluster are incrementally updated with the new samples:
\begin{align}
\tensor{S}_{c^*} &\leftarrow \tensor{S}_{c^*} + \sum_{i\in\mathcal{I}_{c^*}} (\tensor{x}_i-\tensor{\mu}_{c^*})^{\odot 2}, \label{eq_nonnormalized_covariance_uprade}\\
\tensor{\mu}_{c^*} &\leftarrow
\frac{n_{c^*}\,\tensor{\mu}_{c^*} + \sum_{i\in\mathcal{I}_{c^*}} \tensor{x}_i}{n_{c^*}+|\mathcal{I}_{c^*}|}, \label{eq_center_update}\\
n_{c^*} &\leftarrow n_{c^*} + |\mathcal{I}_{c^*}|. \label{eq_count_update}
\end{align}
where {$\odot$ denotes the Hadamard (element-wise) product, so
$(\tensor{x}_i-\tensor{\mu}_{c^*})^{\odot 2}$ is the element-wise square,
$|\mathcal{I}_{c^*}|$ is the cardinality of $\mathcal{I}_{c^*}$, i.e.\ the number of
samples assigned to cluster $c^*$, and} we store the diagonal scatter matrix
$\tensor{S}_c$ for efficient updates and compute $\tensor{\Sigma}_c=\tensor{S}_c/n_c$
when needed.
 The update has
$\mathcal{O}(\vardim)$ complexity for diagonal $\tensor{\Sigma}_c$. 
\subsubsection{Cluster adding}
Let $c^*(i)=\arg\min_{c\in\{1,\dots,C\}} d_{ci}^2$. Samples that are not represented by any existing micro-cluster are indexed as:
\begin{equation}
\mathcal{I}_{\mathrm{+}} = \left\{ i \in \mathcal{I}\ \middle|\  u_{c^*(i),i} \le \exp\!\left(-\kappa_\sigma^2/\vardim\right) \right\}.
\end{equation}
For each $i\in\mathcal{I}_{\mathrm{+}}$, a new micro-cluster is initialized as:
\begin{equation}
\mathcal{C} \leftarrow \mathcal{C} \cup \left\{ \big( \tensor{\Sigma}=
\tensor{\Sigma}_0,\;\tensor{\mu}=\tensor{x}_i,\; n=1\big) \right\}, \qquad i\in\mathcal{I}_{\mathrm{+}}.\label{eq_new_cluster}
\end{equation}
The initial covariance is initialized as: 
\begin{equation}
\tensor{\Sigma}_0
=
\mathrm{diag}\!\left(\tensor{\sigma}^2/\kappa_r\right),\label{eq_cluster_initialization}
\end{equation}
where $\tensor{\sigma}^2\in\mathbb{R}^{\vardim}$ is the incrementally estimated per-dimension data variance and $\kappa_r\in\mathbb{R}^+$ is a quantization parameter.
\subsubsection{Merging mechanism}
\label{subsec:merging_mech}
\par To account for the separability criterion in
Eq.~\eqref{eq_separability_postmerge}, overlapping clusters are merged.
For two clusters $\mathcal{C}_p$ and $\mathcal{C}_q$, the overlap score is defined as:
\begin{align}
{\tensor{\Sigma}_{pq}} &{= \frac{\tensor{S}_p + \tensor{S}_q}{n_p+n_q},} \label{eq_merging_sigma}\\
{d^2_{pq}} &{= (\tensor{\mu}_p-\tensor{\mu}_q)^{\top}\tensor{\Sigma}_{pq}^{-1}(\tensor{\mu}_p-\tensor{\mu}_q),} \label{eq_merging_distance}\\
{\kappa_{pq}} &{= 1 + \frac{n_p n_q}{(n_p+n_q)^2}\, d^2_{pq}.} \label{eq_merging_measure}
\end{align}
Smaller values of $\kappa_{pq}\in(1,\infty)$ correspond to stronger cluster overlap,
with $\kappa_{pq}=1$ when ${\tensor{\mu}_p=\tensor{\mu}_q}$. 
When $\tensor{\Sigma}_{pq}$ is diagonal, the evaluation of $\kappa_{pq}$
has complexity $\mathcal{O}(\vardim)$. A derivation is provided in Appendix~\ref{app:kappa_derivation}.
\par Exhaustive evaluation of all cluster pairs $\{p,q\}$ scales as $\mathcal{O}(C^{2}\vardim)$ and can become prohibitive in high dimensional problems~\cite{Ozbot_Ozawa_skrjanc_2025}.
We therefore restrict evaluation to a pruned $K_\kappa$-nearest-neighbor (kNN) graph constructed on the micro-cluster centers:
\begin{equation}
E_\kappa
=
\Big\{\{p,q\}\ \Big|\ q\in \mathrm{kNN}(\tensor{\mu}_p,K_\kappa),\ p\in \mathrm{kNN}(\tensor{\mu}_q,K_\kappa),\ p\neq q\Big\},
\label{eq_kappa_graph}
\end{equation}
where $\mathrm{kNN}(\tensor{\mu}_p,K_\kappa)$ returns the indices of the $K_\kappa$ closest centers to $\tensor{\mu}_p$ under Euclidean distance, and only mutual kNN pairs are retained to form an undirected sparse candidate graph. In the worst case, this results in at most $C K_\kappa/2$ undirected edges and restricts subsequent pairwise computations to these candidates.
\par Merging is performed greedily on the pruned graph $E_\kappa$ using nearest neighbors. Each cluster $p$ selects the neighbor $q$ with minimal $\kappa_{pq}$ among those satisfying $\kappa_{pq} \le \kappa_{{m}}$. The pair $\{p,q\}$ is merged as~\cite{Ozbot_Ozawa_Skrjanc_2024, skrjanc_2020}:
\begin{align} 
n_{pq} &= n_p + n_q, \label{eq_merging_samples}\\ 
\tensor{\mu}_{pq} &= \frac{n_p\tensor{\mu}_p + n_q\tensor{\mu}_q}{n_{pq}}, \label{eq_merging_centers}\\ 
\tensor{S}_{pq} &= \tensor{S}_p + \tensor{S}_q + \frac{n_p n_q}{n_{pq}}(\tensor{\mu}_p-\tensor{\mu}_q)^{\odot 2}.  \label{eq_merging_spread}
\end{align}
\par The complete evolving clustering in mini-batches is presented in Algorithm \ref{alg_evolving_clustering}.

\begin{algorithm}[!t]
\caption{Evolving Micro-Cluster Learning (mini-batch, parallel)}
\label{alg_evolving_clustering}
\begin{algorithmic}[1]
\Statex \textbf{Input:} Mini-batch $\tensor{X}_b$, micro-clusters $\mathcal{C}$,
$\kappa_\sigma$, $\kappa_m$, $\kappa_r$
\Statex \textbf{Output:} Updated micro-clusters $\mathcal{C}$

\vspace{0.5mm}
\Statex \textbf{Membership computation}
\State Compute Mahalanobis distances $d^2_{ci}$. \Comment{\gray{Eq.~\eqref{eq_mahalanobis_distance}}}
\State Compute memberships $u_{ci}$. \Comment{\gray{Eq.~\eqref{eq_membership}}}

\vspace{0.5mm}
\Statex \textbf{Incremental clustering (representation)}
\State Determine assignment sets $\mathcal{I}_c$. \Comment{\gray{Eq.~\eqref{eq_incremented_samples}}}
\State Update $\{n_c,\tensor{\mu}_c,S_c\}$ simultaneously. \Comment{\gray{Eqs.~\eqref{eq_nonnormalized_covariance_uprade}--\eqref{eq_count_update}}}

\vspace{0.5mm}
\Statex \textbf{Cluster adding (representation failure)}
\State Initialize new micro-clusters. \Comment{\gray{Eqs.~\eqref{eq_new_cluster}, \eqref{eq_cluster_initialization}}}

\vspace{0.5mm}
\Statex \textbf{Merging mechanism (separability)}
\State Construct candidate pairs $E_\kappa$. \Comment{\gray{Eq.~\eqref{eq_kappa_graph}}}
\State Compute overlap scores $\kappa_{pq}$. \Comment{\gray{Eqs.~{\eqref{eq_merging_sigma}--\eqref{eq_merging_measure}}}}
\State Merge pairs satisfying $\kappa_{pq}\le\kappa_m$. \Comment{\gray{Eqs.~\eqref{eq_merging_samples}--\eqref{eq_merging_spread}}}
\end{algorithmic}
\end{algorithm}
\subsection{Density Peak Clustering}
\par Evolving micro-cluster learning can create too many micro-clusters, so $C$ is often larger than the true number of clusters in the data. We adapt DPC~\cite{Rodriguez_Laio_2014} to select $P$ representative peaks from the micro-clusters $\mathcal{C}$. DPC assumes that cluster centers exhibit both high local density and strong separation from other higher-density micro-clusters. For each micro-cluster $i\in\{1,\dots,C\}$, we compute its density $\rho_i$ and its separation~\cite{Rodriguez_Laio_2014}:
\begin{align}
\delta_i &=  \min_{j:\,\rho_j>\rho_i} \lVert \tensor{\mu}_i-\tensor{\mu}_j\rVert_2^2,
\label{eq_dpc_distance_higher}\\
\gamma_i &= \rho_i\,\delta_i,
\label{eq_dpc_selection_criteria}
\end{align}
where $\delta_i$ is the squared Euclidean distance from $\tensor{\mu}_i$ to the nearest micro-cluster center of higher density. We select the $P$ peaks with the largest $\gamma_c$ and denote the corresponding peak micro-clusters by $\mathcal{P}\subseteq\mathcal{C}${, as in Algorithm~\ref{alg_dpc_kappa_peaks}}.
\par DPC applied to micro-clusters $\mathcal{C}$ rather than to individual samples reduces the computational cost from $\mathcal{O}(N^2)$ to $\mathcal{O}(NC)+\mathcal{O}(C^2)$ under the assumption $N \gg C$. Standard DPC defines the local density $\rho_i$ from pairwise distances using either a cutoff $d_c$ or a $k$NN-based kernel. The overlap graph $E_\kappa$ provides a sparse, scale-controlled neighborhood for each micro-cluster, so the density can be estimated from local center-to-center distances only.
We define the micro-cluster density as the accumulated mass within the $E_\kappa$ neighborhood:
\begin{equation}\label{eq_dpc_density}
\rho_i = n_i + \sum\nolimits_{\substack{\{i,j\}\in E_\kappa\\ n_j{>}n_i}} n_j\,\exp(-d_{ij}^2{/d}),
\end{equation}
where $d_{ij}^2$ is computed as in Eq.~(\ref{eq_merging_distance}) {and divided by the dimension $d$ to combat the curse of dimensionality}. The accumulation is distance-weighted and asymmetric ({$n_j>n_i$}), so that {intermediate micro-clusters are lifted toward the dominant one in each region and obtain a small $\delta_i$, leaving a single peak per dense region even when cluster sizes are highly imbalanced}. {This addresses the main failure mode of assigning too many peaks to a single cluster under imbalanced data.}

\par The non-peak micro-clusters are assigned by multi-source shortest-path propagation from the selected peaks over $E_\kappa$, rather than by nearest-higher assignment as in classical DPC. After selecting the $P$ peaks $\mathcal{P}\subseteq\mathcal{C}$, each micro-cluster $i$ is assigned to the peak that minimizes the shortest-path cost:
\begin{align}
\ell(i)
&=
\arg\min_{\;p:\,\mathcal{C}_p\in\mathcal{P}} d_{E_\kappa}^2(p,i),
\label{eq_assign_argmin}\\
d_{E_\kappa}^2(p,i)
&=
\min_{\pi \in \Pi(p,i)} \sum_{\{u,v\}\in\pi} d_{uv}^2{/d},
\label{eq_path_cost}
\end{align}
where $\Pi(p,i)$ denotes the set of paths from node $p$ to $i$ in the graph $(\{1,\dots,C\},E_\kappa)$  and the edge cost $d_{uv}^2{/d}$ is computed as in Eq.~(\ref{eq_merging_distance}). Disconnected components that contain no peak are assigned to the nearest already-assigned micro-cluster by Euclidean distance between centers. We denote the assigned groups by $\mathcal{G}_p=\{\,c\mid \ell(c)=p\,\}$ for each peak $\mathcal{C}_p\in\mathcal{P}$, and the grouping variant as eFDPC-$\mathcal{G}$ (Algorithm~\ref{alg_grouping_dijkstra}).

\begin{algorithm}[!t]
\caption{Peak selection on micro-clusters}
\label{alg_dpc_kappa_peaks}
\begin{algorithmic}[1]
\Statex \textbf{Input:} Micro-clusters $\mathcal{C}$, overlap graph $E_\kappa$, number of peaks $P$
\Statex \textbf{Output:} Peak set $\mathcal{P}\subseteq\mathcal{C}$

\vspace{0.5mm}
\State Compute density $\rho_c$. \Comment{\gray{Eq.~\eqref{eq_dpc_density}}}
\State Compute distance to nearest higher $\delta_c$. \Comment{\gray{Eq.~\eqref{eq_dpc_distance_higher}}}
\State Compute peak score $\gamma_c$. \Comment{\gray{Eq.~\eqref{eq_dpc_selection_criteria}}}
\State Select $\mathcal{P}$ as the $P$ micro-clusters with the largest values of $\gamma_c$.
\end{algorithmic}
\end{algorithm}

\begin{algorithm}[!t]
\caption{Micro-cluster assignment (multi-source geodesic propagation)}
\label{alg_grouping_dijkstra}
\begin{algorithmic}[1]
\Statex \textbf{Input:} Micro-clusters $\mathcal{C}$, overlap graph $E_\kappa$, peaks $\mathcal{P}\subseteq\mathcal{C}$
\Statex \textbf{Output:} Peak label $\ell(c)\in\{\,p \mid \mathcal{C}_p\in\mathcal{P}\,\}$ for all $c\in\{1,\dots,C\}$

\vspace{0.5mm}
\State Initialize $\ell(c)\leftarrow -1$ and $d(c)\leftarrow +\infty$ for all $c$.
\For{each peak $\mathcal{C}_p \in \mathcal{P}$}
\State $d(p)\leftarrow 0$ and $\ell(p)\leftarrow p$.
\EndFor

\vspace{0.5mm}
\State Set the edge cost on $E_\kappa$ to $d_{uv}^2{/d}$. \Comment{Eq.~(\ref{eq_merging_distance})}
\State $\ell(\cdot)\gets$ multi-source Dijkstra labels from $\mathcal{P}$ on $E_\kappa$. \Comment{\gray{Eqs.~\eqref{eq_assign_argmin}--\eqref{eq_path_cost}}}

\Statex \textbf{Fallback (assign disconnected components)}
\For{each $c$ with $\ell(c)=-1$}
\State $\ell(c)\gets \ell\!\left(\arg\min_{j:\,\ell(j)\neq -1}\|\tensor{\mu}_c-\tensor{\mu}_j\|_2^2\right)$.
\EndFor
\end{algorithmic}
\end{algorithm}

\subsection{Federated Learning}
\label{subsec:federated_learning}
\par In horizontal federated learning, the data are partitioned across $K\in\mathbb{Z}^+$ clients, where client $k$ holds $\tensor{X}^{(k)}\in\mathbb{R}^{N^{(k)}\times\vardim}$ and $N=\sum_{k=1}^{K}N^{(k)}$. Each client learns micro-clusters $\mathcal{C}^{(k)}$ from its local data and then selects a subset of peaks $\mathcal{P}^{(k)}$ from $\mathcal{C}^{(k)}$. We assume an honest-but-curious server, \emph{i.e.}, the server follows the algorithm exactly but may attempt to extract additional information from received messages~\cite{Hardy_Henecka_Ivey_Law_Nock_Patrini_Smith_Thorne_2017, Kanamori_Abe_2022}. We also assume that the server needs to have access to the global model, so privacy protection focuses on preventing exposure of raw samples and ensuring unlinkability of client contributions. To avoid revealing which client produced which peaks, we use an additional shuffler-based aggregation layer between the clients and the server~\cite{Bittau_Erlingsson_Maniatis_Mironov_Raghunathan_Lie_Rudominer_Kode_Tinnes_Seefeld_2017}. {Unlinkability assumes that the shufflers do not collude with the server. For each transmitted peak, the client creates new additive shares and a new random identifier that is used only to match those shares during reconstruction. Therefore, the server can reconstruct the peak statistics from the shuffled shares, but does not know which client produced each reconstructed peak.}

\par Each transmitted peak statistic consists of the center $\tensor{\mu}^{(k)}_c$ and the min--max normalized {support} $\tilde n^{(k)}_c$, obtained by scaling the local peak {support} $\{n^{(k)}_c\}_{c=1}^{C^{(k)}}$ to the unit interval $[0,1]$. For each peak $c$, the client generates $S-1$ random additive shares:
\begin{equation}\label{eq:shuffler-share-random}
(\tensor{\mu}^{(k,s)}_{c},\tilde n^{(k,s)}_{c}) \sim \mathcal{D}_{\mu,\tilde n},
\qquad s=1,\dots,S-1,
\end{equation}
and sets the last share so that the $S$ shares sum to the true values:
\begin{equation}\label{eq:shuffler-last-share}
\tensor{\mu}^{(k,S)}_{c} = \tensor{\mu}^{(k)}_{c} - \sum_{s=1}^{S-1}\tensor{\mu}^{(k,s)}_{c},
\qquad \tilde n^{(k,S)}_{c} = \tilde n^{(k)}_{c} - \sum_{s=1}^{S-1}\tilde n^{(k,s)}_{c}.
\end{equation}

\par To enable reconstruction at the server, a public identifier is used to match shares of the same peak within a round. For each peak $c$, the client attaches a {random} per-peak identifier $t_c$ and sends share-messages to each shuffler $s\in{1,\dots,S}$. We denote the message set sent from client $k$ to shuffler $s$ and the unordered set forwarded by shuffler $s$ as
\begin{align}
\mathcal{M}^{(k,s)}
&=
\{(t_c,\tensor{\mu}^{(k,s)}_c,\tilde n^{(k,s)}_c)\mid \mathcal{C}^{(k)}_c\in\mathcal{P}^{(k)}\},
\label{eq:shuffler_messages_client}\\
\mathcal{M}^{(s)}
&=
\bigcup_{k=1}^{K}\mathcal{M}^{(k,s)}.
\label{eq:shuffler_messages_shuffler}
\end{align}
The shuffler forwards the received messages in a uniformly random order. The server reconstructs each peak by grouping messages with the same identifier $t$ and summing the corresponding shares:
\begin{equation}\label{eq:shuffler_reconstruct_sum}
\tensor{\mu}_t=\sum_{s=1}^{S}\tensor{\mu}^{(s)}_t,\qquad
\tilde n_t=\sum_{s=1}^{S}\tilde n^{(s)}_t.
\end{equation}
The shuffling step breaks the association between reconstructed peaks and their client of origin at the server. Each shuffler receives only one additive share per peak, which is insufficient to reconstruct the peaks.

\par The server still requires a diagonal covariance to evaluate the overlap score in Eqs.~\eqref{eq_merging_sigma}{--\eqref{eq_merging_measure}} and the edge costs used in DPC. The server therefore assigns each reconstructed peak a diagonal covariance proxy using Eq. \eqref{eq_cluster_initialization}. After reconstructing the peak centers and their normalized supports, the server estimates a per-dimension spread across the reconstructed peaks and scales it by $\kappa_r$ to obtain a diagonal covariance for each peak. Each reconstructed peak is then treated as a diagonal-covariance micro-cluster and processed by the same overlap and DPC routines as at the client{-}side. eFDPC is presented in Algorithm~\ref{alg_federated_eFDPC}. {The algorithm is written for the general multi-round federated setting, and the while loop denotes standard iterative communication. All our experiments use a single round, which is sufficient. eFDPC therefore operates one-shot in practice while still supporting multi-round refinement.}

\begin{algorithm}[!t]
\caption{eFDPC}
\label{alg_federated_eFDPC}
\begin{algorithmic}[1]
\State \textbf{Parameters:} $\kappa_m, \kappa_r$
\State \textbf{Initialization:} $\mathcal{P} \leftarrow \emptyset$

\While{federated learning}

\vspace{0.5mm}
\Statex \quad~ \textbf{Local learning (client $k=1,\dots,K$):}
\For{$k \in \{1,2,\dots,K\}$}
    \State $\mathcal{C}^{(k)} \gets$ Evolving micro-cluster learning on local data
    \Comment{\gray{Algorithm~\ref{alg_evolving_clustering}}}
    \State $\mathcal{P}^{(k)} \gets$ Peak selection on $\mathcal{C}^{(k)}$
    \Comment{\gray{Algorithm~\ref{alg_dpc_kappa_peaks}}}
\EndFor
\vspace{0.5mm}
\Statex \quad~\textbf{Private communication:}
\State Form $\mathcal{M}^{(k,s)}$ from $\mathcal{P}^{(k)}$ and send to each shuffler $s$.
\Comment{\gray{Eq.~\eqref{eq:shuffler_messages_client}}}
\State Shufflers compute $\mathcal{M}^{(s)}$ and forward to server.
\Comment{\gray{Eq.~\eqref{eq:shuffler_messages_shuffler}}}
\State Reconstruct $\mathcal{C}_g$ by summing shares with the same identifier $t$. 
\Comment{\gray{Eq.~\eqref{eq:shuffler_reconstruct_sum}}}%

\vspace{0.5mm}
\Statex \quad~\textbf{Server refinement:}
\State $\mathcal{C}_g \gets$ Merging mechanism on $\mathcal{C}_g$
\Comment{\gray{Section~\ref{subsec:merging_mech}}}
\State $\mathcal{P}_g\gets$ Peak selection on $\mathcal{C}_g$
\Comment{\gray{Algorithm~\ref{alg_dpc_kappa_peaks}}}
\State (Optional) $\mathcal{C}_g \gets$ Propagate labels from $\mathcal{P}_g$
\Comment{\gray{Algorithm~\ref{alg_grouping_dijkstra}}}%

\vspace{0.5mm}
\Statex \quad~\textbf{ Synchronization:}
\State Broadcast $\mathcal{P}_g$ to all clients

\vspace{0.5mm}
\EndWhile
\end{algorithmic}
\end{algorithm}

\subsection{{Computational cost}}
{Table~\ref{tab_cost} summarizes the computational and memory cost of eFDPC. Since the micro-clusters use diagonal covariance matrices, the main operations scale linearly with the dimension $d$, and each micro-cluster requires $\mathcal{O}(d)$ memory. For client $k$, the main learning cost is $\mathcal{O}(N^{(k)}C^{(k)}d)$, while local peak selection costs $\mathcal{O}((C^{(k)})^2d)$. The local model requires $\mathcal{O}(C^{(k)}d)$ memory. The shuffler layer adds $\mathcal{O}(SPd)$ operations for generating additive shares. Since eFDPC is one-shot, this is done only once. Each client sends $P$ peak summaries as $S$ additive shares in a single communication round. The reported runtimes include only local computation and exclude communication.}

\begin{table}[!t]
\centering
\caption{{Computational and memory complexity of eFDPC.}}
\label{tab_cost}
\begin{tabular}{llcc}
\toprule
{Stage} &{Operation} &{Time} &{Memory} \\
\midrule
{\multirow{4}{*}{Client}}
 &{Micro-cluster update (per batch)} &{$\mathcal{O}(N_b C d)$} &{\multirow{4}{*}{$\mathcal{O}((C{+}N_b)d)$}} \\
 &{Micro-cluster update (full pass)} &{$\mathcal{O}(N C d)$}   & \\
 &{Local DPC peak selection}        &{$\mathcal{O}(C^2 d)$}  & \\
 &{Share generation}                &{$\mathcal{O}(SPd)$}    & \\
\midrule
{Communication} &{Per client, single round} &{$\mathcal{O}(Pd)$} &{$\mathcal{O}(Pd)$} \\
\midrule
{\multirow{2}{*}{Server}}
 &{Peak reconstruction}            &{$\mathcal{O}(SKPd)$}    &{\multirow{2}{*}{$\mathcal{O}(KPd)$}} \\
 &{Global DPC \& label propagation} &{$\mathcal{O}((KP)^2 d)$} & \\
\bottomrule
\end{tabular}
\end{table}

\section{{Experimentation}}\label{sec:experimentation}
\par The proposed eFDPC method was evaluated on a collection of synthetic and real-world clustering benchmarks. 
The synthetic datasets in Table~\ref{tab:clustering_synth} include Compound~\cite{compound_Zahn_1971},
D31 and R15~\cite{D31_R15_Veenman_Reinders_Backer_2002},
S3 and S4~\cite{Ssets}, Flame~\cite{Fu_Medico_2007}
Aggregation~\cite{aggregations_Gionis_Mannila_Tsaparas_2007},
Jain~\cite{Jain_Law_2005},
Pathbased,
and the Variations datasets Asymmetric, Skewed, Unbalance, and Overlap~\cite{variations_Rezaei_Franti_2020}.
These datasets are commonly used to evaluate clustering performance under controlled geometric conditions in low-dimensional space.
Real-world datasets in Table~\ref{tab:clustering_real} include tabular benchmarks from UCI and related sources~\cite{Kelly_Longjohn_Nottingham_UCI}, namely Dermatology, Seeds, WDBC, Heart, Iris, Leaf, Libras, Texture, Thyroid, and Wine, as well as higher-dimensional datasets including MNIST and KDD99.
These datasets cover a wide range of dimensionalities and sample sizes, from small tabular problems to large-scale high-dimensional settings.   For baseline comparison, we used recent federated clustering approaches with publicly available implementations, namely FKDC~\cite{Wang_Pang_Wang_Pedrycz_2025}, $k$-FED~\cite{Dennis_Li_Smith_2021}, MUFC~\cite{Pan_Sima_Prakash_Rana_Milenkovic_2022}, FedFCM~\cite{Stallmann_Wilbik_2022}, and LLF-CM and LLF-FCM~\cite{Corcuera_Barcena_Marcelloni_Renda_Bechini_Ducange_2024}.
\par The datasets were partitioned using the horizontal federated setting described in Section~\ref{subsec:federated_learning}. Performance was measured using clustering accuracy (ACC), Adjusted Rand Index (ARI), Normalized Mutual Information (NMI), and total wall-clock runtime in seconds (local computation only, excluding communication). For the non-IID setting, the dataset was split across $K$ clients using a class-wise Dirichlet partition~\cite{Li_Diao_Chen_He_2021, Hsu_Qi_Brown_2019} with $\alpha=0.9$. In each {run}, samples within each class were randomly permuted and assigned to clients according to {these proportions}, with {a nonzero number of samples enforced for each client}. The same client partition was reused across all compared methods within {that run}. All experiments were repeated for 20 independent non-IID partitions, and the mean measures {were averaged across them}. In all experiments, the true number of clusters is used as the number of peaks $P$ for eFDPC and as the cluster count for all baselines, so that all methods are given the true number of clusters as input, which in practice is unknown. The eFDPC parameters were set using simple heuristic rules: $\kappa_\sigma = d$ since squared Mahalanobis distances satisfy $d_M^2 \sim$ {$\kappa_r\,\chi^2(d)$, so the acceptance radius $\kappa_\sigma^2/\kappa_r$ matches the expected chi-squared distance $\mathbb{E}[\chi^2_d] = d$ (with $\kappa_r = d$)}; the overlap-based merging threshold used the normalized chi-squared rule $\kappa_m = 1+\frac{1}{4}\frac{\chi^2_d{}^{-1}(p)}{d}$ with the chi-squared median $p=0.5$; $N_b=\sqrt{N}$ followed the common heuristic $C\approx\sqrt{N}$; $K_\kappa=\sqrt{C}$; and $\kappa_r=\max(4P,d)$ based on the empirical observation that $P$ dominates in low dimensions, whereas $d$ dominates in high dimensions; all parameters were fixed across all datasets and were not tuned per dataset.

\par
\begin{figure}[!t]
    \centering
    \begin{minipage}[t]{0.48\linewidth}\centering
        \includegraphics[width=\linewidth]{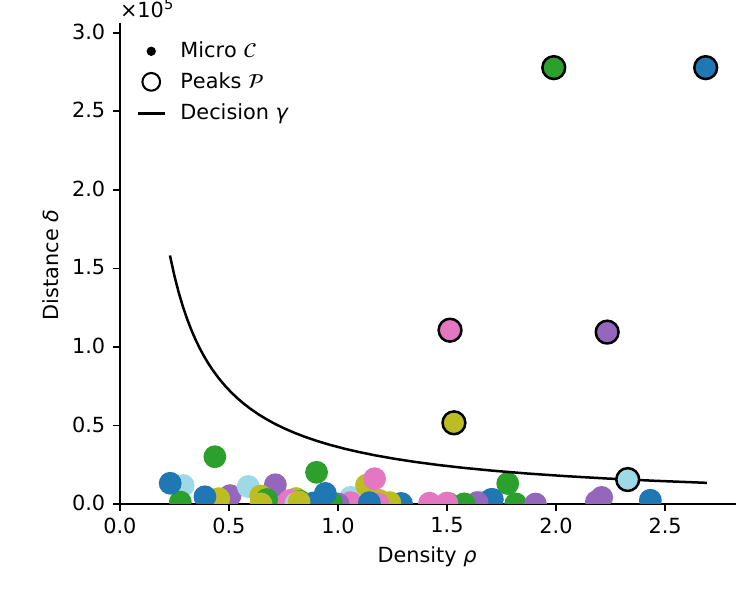}\\[-10pt]{\footnotesize{(a) Skewed, decision graph}}
    \end{minipage}\hfill
    \begin{minipage}[t]{0.48\linewidth}\centering
        \includegraphics[width=\linewidth]{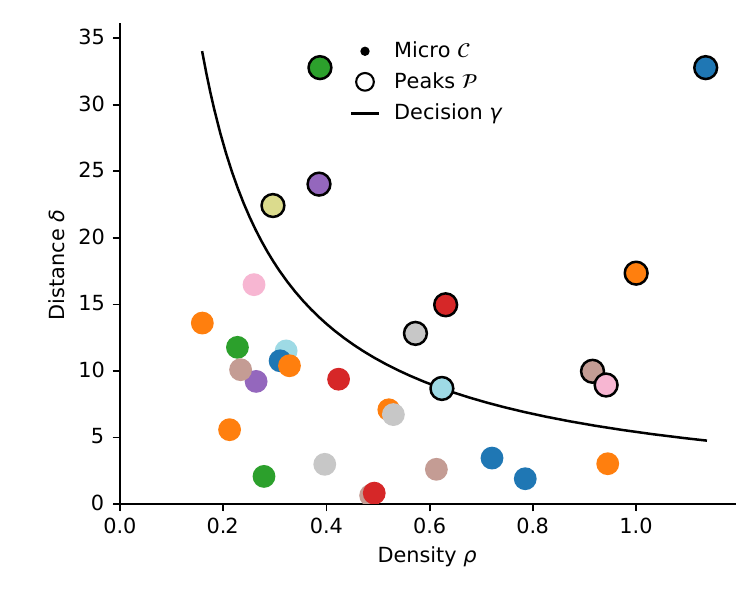}\\[-10pt]{\footnotesize{(b) MNIST, decision graph}}
    \end{minipage}

    \begin{minipage}[t]{0.48\linewidth}\centering
        \includegraphics[width=\linewidth]{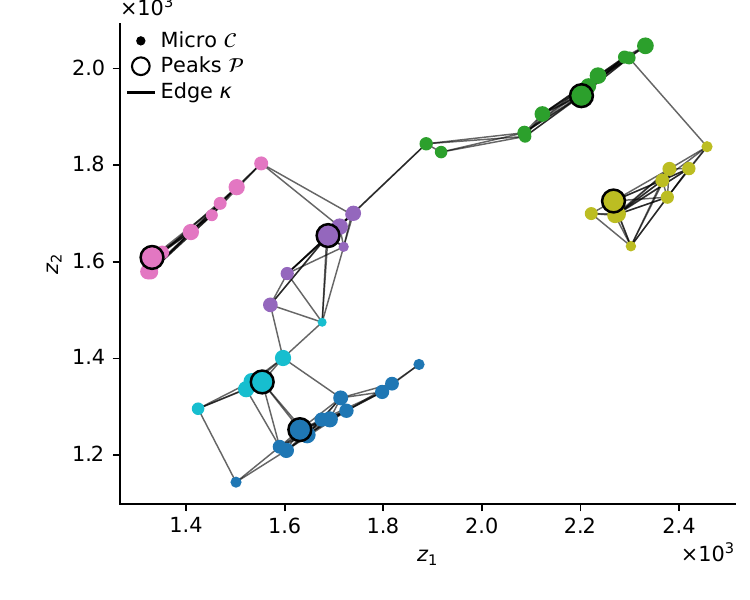}\\[-10pt]{\footnotesize{(c) Skewed, grouping}}
    \end{minipage}\hfill
    \begin{minipage}[t]{0.48\linewidth}\centering
        \includegraphics[width=\linewidth]{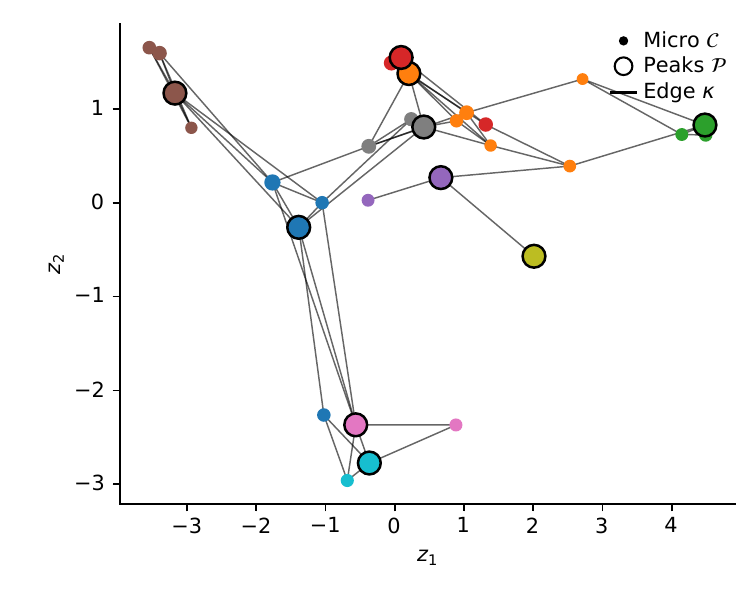}\\[-10pt]{\footnotesize{(d) MNIST, grouping}}
    \end{minipage}

    \begin{minipage}[t]{0.48\linewidth}\centering
        \includegraphics[width=\linewidth]{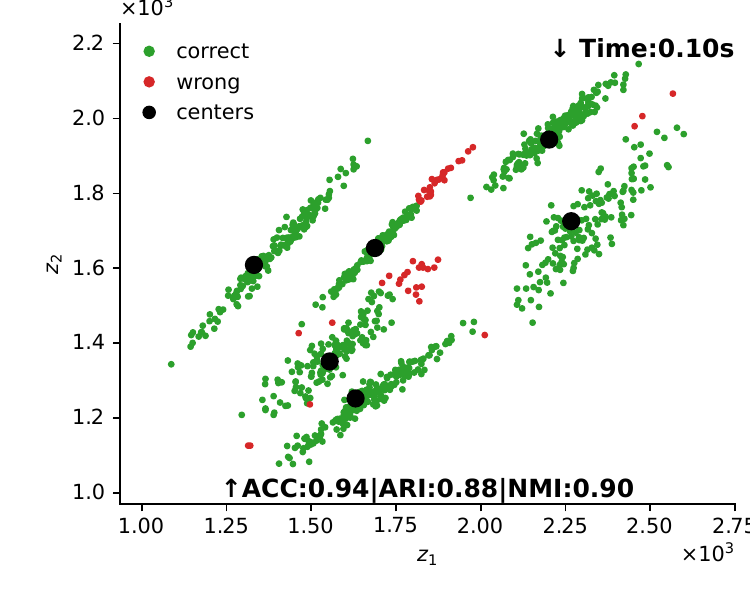}\\[-10pt]{\footnotesize{(e) Skewed, accuracy}}
    \end{minipage}\hfill
    \begin{minipage}[t]{0.48\linewidth}\centering
        \includegraphics[width=\linewidth]{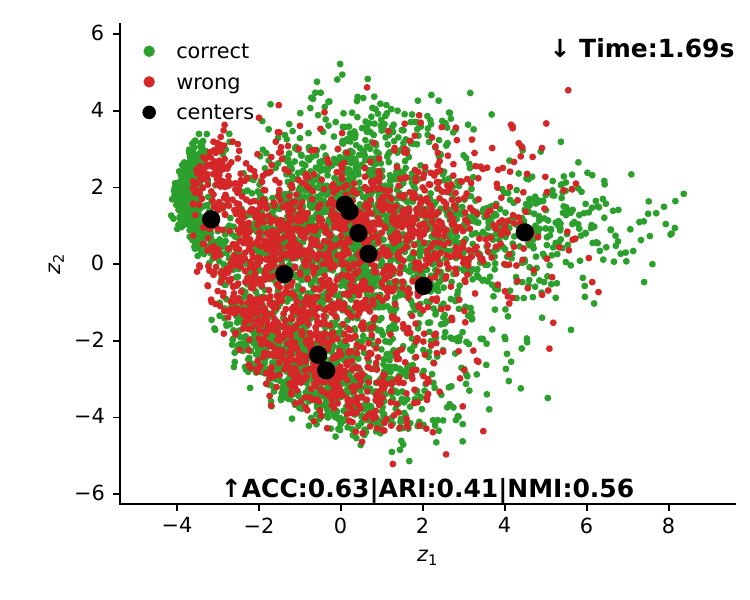}\\[-10pt]{\footnotesize{(f) MNIST, accuracy}}
    \end{minipage}

    \caption{Comparison of eFDPC-$\mathcal{G}$ behavior on Skewed ($K=10$, left column) and
    MNIST ($K=3$, right column). From top to bottom, rows depict the decision graph (a,b),
    cluster grouping (c,d), and clustering accuracy (e,f) at the server{-}side. The
    evaluation is performed using the propagated labels on the shared peaks.}
    \label{fig_decision_grouping_skewed_mnist}
\end{figure}

 {Figure}~\ref{fig_decision_grouping_skewed_mnist} illustrates server-side peak selection and label propagation for Skewed ($K=10$ clients) and MNIST ($K=3$ clients). Each client transmits $P$ local peaks, and the server selects $P$ global peaks from the reconstructed pool. Increasing the client{-}count improves the aggregated connectivity among reconstructed candidates, making the propagation step more informative on Skewed. In contrast, MNIST is more uniform, so peak selection is less clear and less stable. However, label propagation stabilizes the result.

\begin{table}[!t]
  \centering
  \footnotesize
  \caption{Clustering comparison on synthetic datasets ($K=3$, $\alpha=0.9$){.}}\label{tab:clustering_synth}
  \setlength{\tabcolsep}{2pt}
  \begin{tabular}{l|l|llllllll}
    \toprule
    Dataset & Metric & eFDPC-$\mathcal{P}$ & eFDPC-$\mathcal{G}$ & FKDC & $k$-FED & MUFC & FedFCM & LLF-FCM & LLF-CM \\
    \midrule
    \multirow{4}{*}{\shortstack[l]{{Aggregation}\\$N=788$\\$\vardim=2$}} & ACC [$10^{-2}$] & 92$\pm$02 & \textbf{92$\pm$02} & 85$\pm$03 & 89$\pm$03 & 88$\pm$03 & 82$\pm$05 & 79$\pm$04 & 80$\pm$05 \\
     & ARI [$10^{-2}$] & 86$\pm$05 & \textbf{86$\pm$04} & 75$\pm$06 & 81$\pm$05 & 82$\pm$05 & 70$\pm$06 & 67$\pm$07 & 68$\pm$07 \\
     & NMI [$10^{-2}$] & 89$\pm$02 & \textbf{90$\pm$02} & 81$\pm$04 & 86$\pm$04 & 85$\pm$03 & 76$\pm$06 & 75$\pm$05 & 76$\pm$05 \\
     & Time [$10^{-3}{\mathrm{s}}$] & 98$\pm$07 & 98$\pm$07 & 94$\pm$13 & 29$\pm$04 & 88$\pm$06 & 27$\pm$10 & 09$\pm$00 & \textbf{02$\pm$00} \\
    \midrule
    \multirow{4}{*}{\shortstack[l]{{Asymmetric}\\$N=1000$\\$\vardim=2$}} & ACC [$10^{-2}$] & \textbf{96$\pm$02} & 95$\pm$03 & 80$\pm$09 & 87$\pm$10 & 94$\pm$05 & 89$\pm$07 & 77$\pm$09 & 76$\pm$10 \\
     & ARI [$10^{-2}$] & \textbf{90$\pm$04} & 90$\pm$05 & 71$\pm$11 & 81$\pm$12 & 88$\pm$06 & 81$\pm$09 & 66$\pm$12 & 66$\pm$11 \\
     & NMI [$10^{-2}$] & 91$\pm$02 & \textbf{91$\pm$03} & 83$\pm$04 & 87$\pm$05 & 90$\pm$03 & 86$\pm$05 & 77$\pm$07 & 77$\pm$07 \\
     & Time [$10^{-2}{\mathrm{s}}$] & 10$\pm$00 & 10$\pm$00 & 09$\pm$01 & 02$\pm$00 & 08$\pm$00 & 00$\pm$00 & 00$\pm$00 & \textbf{00$\pm$00} \\
    \midrule
    \multirow{4}{*}{\shortstack[l]{{Compound}\\$N=399$\\$\vardim=2$}} & ACC [$10^{-2}$] & \textbf{83$\pm$05} & 83$\pm$04 & 79$\pm$03 & 80$\pm$04 & 81$\pm$03 & 75$\pm$02 & 76$\pm$04 & 78$\pm$04 \\
     & ARI [$10^{-2}$] & 76$\pm$04 & \textbf{77$\pm$02} & 74$\pm$02 & 75$\pm$03 & 75$\pm$03 & 70$\pm$03 & 68$\pm$08 & 70$\pm$08 \\
     & NMI [$10^{-2}$] & 79$\pm$04 & \textbf{81$\pm$02} & 75$\pm$02 & 78$\pm$04 & 76$\pm$03 & 72$\pm$03 & 70$\pm$07 & 73$\pm$07 \\
     & Time [$10^{-3}{\mathrm{s}}$] & 68$\pm$03 & 68$\pm$03 & 76$\pm$14 & 27$\pm$03 & 65$\pm$09 & 11$\pm$03 & 04$\pm$00 & \textbf{01$\pm$00} \\
    \midrule
    \multirow{4}{*}{\shortstack[l]{{D31}\\$N=3100$\\$\vardim=2$}} & ACC [$10^{-2}$] & \textbf{97$\pm$00} & 97$\pm$00 & 63$\pm$03 & 82$\pm$03 & 93$\pm$02 & 75$\pm$02 & 70$\pm$03 & 71$\pm$03 \\
     & ARI [$10^{-2}$] & \textbf{94$\pm$00} & 94$\pm$00 & 56$\pm$03 & 79$\pm$03 & 88$\pm$02 & 68$\pm$02 & 62$\pm$03 & 65$\pm$03 \\
     & NMI [$10^{-2}$] & \textbf{96$\pm$00} & 96$\pm$00 & 81$\pm$01 & 91$\pm$01 & 94$\pm$01 & 86$\pm$01 & 84$\pm$01 & 85$\pm$01 \\
     & Time [$10^{-1}{\mathrm{s}}$] & 02$\pm$00 & 02$\pm$00 & 02$\pm$00 & 00$\pm$00 & 05$\pm$00 & 14$\pm$05 & 01$\pm$00 & \textbf{00$\pm$00} \\
    \midrule
    \multirow{4}{*}{\shortstack[l]{{Flame}\\$N=240$\\$\vardim=2$}} & ACC [$10^{-2}$] & 80$\pm$07 & \textbf{82$\pm$09} & 80$\pm$07 & 81$\pm$07 & 78$\pm$08 & 78$\pm$07 & 80$\pm$05 & 77$\pm$06 \\
     & ARI [$10^{-2}$] & 39$\pm$17 & \textbf{45$\pm$26} & 37$\pm$18 & 42$\pm$16 & 33$\pm$18 & 33$\pm$17 & 37$\pm$12 & 32$\pm$16 \\
     & NMI [$10^{-2}$] & 35$\pm$15 & \textbf{43$\pm$21} & 35$\pm$15 & 37$\pm$13 & 32$\pm$15 & 31$\pm$16 & 36$\pm$11 & 33$\pm$12 \\
     & Time [$10^{-3}{\mathrm{s}}$] & 48$\pm$04 & 48$\pm$04 & 54$\pm$09 & 28$\pm$02 & 45$\pm$03 & 02$\pm$00 & 01$\pm$00 & \textbf{00$\pm$00} \\
    \midrule
    \multirow{4}{*}{\shortstack[l]{{Jain}\\$N=373$\\$\vardim=2$}} & ACC [$10^{-2}$] & \textbf{90$\pm$02} & 90$\pm$05 & 85$\pm$08 & 88$\pm$03 & 88$\pm$05 & 84$\pm$03 & 83$\pm$05 & 85$\pm$07 \\
     & ARI [$10^{-2}$] & 64$\pm$08 & \textbf{64$\pm$16} & 49$\pm$25 & 57$\pm$10 & 59$\pm$18 & 47$\pm$08 & 45$\pm$16 & 50$\pm$24 \\
     & NMI [$10^{-2}$] & 56$\pm$04 & \textbf{57$\pm$11} & 45$\pm$20 & 50$\pm$04 & 51$\pm$14 & 47$\pm$05 & 44$\pm$13 & 46$\pm$18 \\
     & Time [$10^{-3}{\mathrm{s}}$] & 61$\pm$03 & 61$\pm$03 & 66$\pm$11 & 27$\pm$03 & 51$\pm$06 & 02$\pm$00 & 01$\pm$00 & \textbf{00$\pm$00} \\
    \midrule
    \multirow{4}{*}{\shortstack[l]{{Overlap}\\$N=1000$\\$\vardim=2$}} & ACC [$10^{-2}$] & \textbf{82$\pm$04} & 80$\pm$04 & 78$\pm$04 & 79$\pm$06 & 76$\pm$05 & 73$\pm$05 & 49$\pm$06 & 22$\pm$07 \\
     & ARI [$10^{-2}$] & \textbf{68$\pm$03} & 64$\pm$04 & 66$\pm$02 & 65$\pm$05 & 59$\pm$04 & 55$\pm$06 & 36$\pm$05 & 04$\pm$08 \\
     & NMI [$10^{-2}$] & 73$\pm$02 & 72$\pm$02 & \textbf{75$\pm$00} & 72$\pm$03 & 68$\pm$03 & 63$\pm$04 & 54$\pm$04 & 08$\pm$15 \\
     & Time [$10^{-2}{\mathrm{s}}$] & 10$\pm$00 & 10$\pm$00 & 09$\pm$01 & 02$\pm$00 & 09$\pm$00 & 03$\pm$01 & 00$\pm$00 & \textbf{00$\pm$00} \\
    \midrule
    \multirow{4}{*}{\shortstack[l]{{R15}\\$N=600$\\$\vardim=2$}} & ACC [$10^{-2}$] & \textbf{99$\pm$00} & 99$\pm$00 & 59$\pm$04 & 88$\pm$05 & 97$\pm$02 & 81$\pm$05 & 73$\pm$05 & 68$\pm$04 \\
     & ARI [$10^{-2}$] & \textbf{98$\pm$00} & 98$\pm$00 & 46$\pm$10 & 86$\pm$05 & 96$\pm$03 & 76$\pm$06 & 65$\pm$05 & 60$\pm$06 \\
     & NMI [$10^{-2}$] & \textbf{99$\pm$00} & 98$\pm$00 & 78$\pm$03 & 94$\pm$02 & 97$\pm$01 & 89$\pm$02 & 85$\pm$02 & 84$\pm$02 \\
     & Time [$10^{-2}{\mathrm{s}}$] & 09$\pm$00 & 09$\pm$00 & 09$\pm$01 & 03$\pm$00 & 10$\pm$00 & 04$\pm$02 & 01$\pm$00 & \textbf{00$\pm$00} \\
    \midrule
    \multirow{4}{*}{\shortstack[l]{{S3}\\$N=5000$\\$\vardim=2$}} & ACC [$10^{-2}$] & \textbf{84$\pm$01} & 84$\pm$01 & 59$\pm$04 & 75$\pm$03 & 78$\pm$02 & 72$\pm$04 & 63$\pm$04 & 60$\pm$04 \\
     & ARI [$10^{-2}$] & \textbf{70$\pm$01} & 70$\pm$01 & 47$\pm$04 & 63$\pm$03 & 64$\pm$02 & 58$\pm$04 & 51$\pm$03 & 49$\pm$04 \\
     & NMI [$10^{-2}$] & 78$\pm$00 & \textbf{78$\pm$00} & 68$\pm$02 & 76$\pm$01 & 76$\pm$01 & 73$\pm$02 & 69$\pm$01 & 69$\pm$02 \\
     & Time [$10^{-2}{\mathrm{s}}$] & 34$\pm$02 & 34$\pm$02 & 21$\pm$01 & 05$\pm$00 & 43$\pm$01 & 85$\pm$29 & 12$\pm$04 & \textbf{03$\pm$01} \\
    \midrule
    \multirow{4}{*}{\shortstack[l]{{S4}\\$N=5000$\\$\vardim=2$}} & ACC [$10^{-2}$] & \textbf{78$\pm$01} & 77$\pm$01 & 56$\pm$04 & 71$\pm$02 & 72$\pm$03 & 68$\pm$03 & 59$\pm$03 & 55$\pm$03 \\
     & ARI [$10^{-2}$] & \textbf{62$\pm$01} & 60$\pm$01 & 42$\pm$03 & 56$\pm$01 & 56$\pm$03 & 52$\pm$03 & 45$\pm$02 & 42$\pm$03 \\
     & NMI [$10^{-2}$] & \textbf{71$\pm$00} & 71$\pm$00 & 61$\pm$01 & 69$\pm$00 & 69$\pm$01 & 67$\pm$01 & 63$\pm$01 & 62$\pm$01 \\
     & Time [$10^{-2}{\mathrm{s}}$] & 30$\pm$03 & 30$\pm$03 & 33$\pm$47 & 06$\pm$00 & 45$\pm$01 & 89$\pm$26 & 13$\pm$09 & \textbf{04$\pm$02} \\
    \midrule
    \multirow{4}{*}{\shortstack[l]{{Skewed}\\$N=1000$\\$\vardim=2$}} & ACC [$10^{-2}$] & 81$\pm$04 & \textbf{81$\pm$05} & 81$\pm$06 & 73$\pm$04 & 73$\pm$07 & 64$\pm$06 & 44$\pm$09 & 22$\pm$07 \\
     & ARI [$10^{-2}$] & 70$\pm$04 & \textbf{71$\pm$07} & 70$\pm$07 & 62$\pm$05 & 60$\pm$08 & 48$\pm$06 & 33$\pm$10 & 05$\pm$09 \\
     & NMI [$10^{-2}$] & 77$\pm$02 & \textbf{79$\pm$04} & 78$\pm$04 & 73$\pm$03 & 71$\pm$05 & 61$\pm$05 & 51$\pm$08 & 10$\pm$16 \\
     & Time [$10^{-2}{\mathrm{s}}$] & 10$\pm$00 & 10$\pm$00 & 10$\pm$01 & 03$\pm$00 & 09$\pm$00 & 02$\pm$00 & 01$\pm$00 & \textbf{00$\pm$00} \\
    \midrule
    \multirow{4}{*}{\shortstack[l]{{Unbalance}\\$N=6500$\\$\vardim=2$}} & ACC [$10^{-2}$] & 99$\pm$00 & \textbf{99$\pm$00} & 99$\pm$00 & 98$\pm$01 & 98$\pm$00 & 72$\pm$10 & 86$\pm$10 & 35$\pm$11 \\
     & ARI [$10^{-2}$] & 99$\pm$00 & \textbf{99$\pm$00} & 99$\pm$01 & 97$\pm$01 & 98$\pm$01 & 58$\pm$14 & 78$\pm$15 & 07$\pm$17 \\
     & NMI [$10^{-2}$] & 99$\pm$00 & \textbf{99$\pm$00} & 98$\pm$01 & 96$\pm$02 & 97$\pm$01 & 68$\pm$08 & 80$\pm$09 & 10$\pm$22 \\
     & Time [$10^{-2}{\mathrm{s}}$] & 31$\pm$00 & 31$\pm$00 & 18$\pm$01 & 04$\pm$00 & 25$\pm$01 & \textbf{00$\pm$00} & 08$\pm$00 & 02$\pm$00 \\
    \bottomrule
  \end{tabular}
\end{table}

\begin{table}[!t]
  \centering
  \footnotesize
  \caption{Clustering comparison on real-world datasets ($K=3$, $\alpha=0.9$){.}}\label{tab:clustering_real}
  \setlength{\tabcolsep}{2pt}
  \begin{tabular}{l|l|llllllll}
    \toprule
    Dataset & Metric & eFDPC-$\mathcal{P}$ & eFDPC-$\mathcal{G}$ & FKDC & $k$-FED & MUFC & FedFCM & LLF-FCM & LLF-CM \\
    \midrule
    \multirow{4}{*}{\shortstack[l]{{Dermatology}\\$N=358$\\$\vardim=34$}} & ACC [$10^{-2}$] & \textbf{88$\pm$04} & 88$\pm$03 & 68$\pm$04 & 79$\pm$08 & 80$\pm$04 & 74$\pm$05 & 71$\pm$06 & 66$\pm$08 \\
     & ARI [$10^{-2}$] & \textbf{87$\pm$03} & 86$\pm$04 & 57$\pm$07 & 75$\pm$11 & 72$\pm$07 & 67$\pm$07 & 61$\pm$11 & 51$\pm$17 \\
     & NMI [$10^{-2}$] & \textbf{91$\pm$01} & 91$\pm$02 & 73$\pm$04 & 85$\pm$07 & 78$\pm$04 & 76$\pm$05 & 67$\pm$10 & 65$\pm$10 \\
     & Time [$10^{-3}{\mathrm{s}}$] & 48$\pm$03 & 48$\pm$03 & 76$\pm$08 & 29$\pm$03 & 59$\pm$07 & 19$\pm$07 & 04$\pm$01 & \textbf{01$\pm$00} \\
    \midrule
    \multirow{4}{*}{\shortstack[l]{{Heart}\\$N=270$\\$\vardim=13$}} & ACC [$10^{-2}$] & 78$\pm$05 & \textbf{79$\pm$04} & 76$\pm$05 & 75$\pm$07 & 71$\pm$09 & 76$\pm$05 & 73$\pm$07 & 70$\pm$07 \\
     & ARI [$10^{-2}$] & 32$\pm$10 & \textbf{34$\pm$08} & 29$\pm$09 & 27$\pm$13 & 21$\pm$14 & 28$\pm$09 & 24$\pm$11 & 18$\pm$11 \\
     & NMI [$10^{-2}$] & 26$\pm$06 & \textbf{27$\pm$05} & 22$\pm$07 & 21$\pm$09 & 17$\pm$10 & 22$\pm$07 & 19$\pm$09 & 16$\pm$09 \\
     & Time [$10^{-3}{\mathrm{s}}$] & 42$\pm$02 & 42$\pm$02 & 60$\pm$08 & 28$\pm$03 & 48$\pm$05 & 05$\pm$02 & 01$\pm$00 & \textbf{00$\pm$00} \\
    \midrule
    \multirow{4}{*}{\shortstack[l]{{Iris}\\$N=150$\\$\vardim=4$}} & ACC [$10^{-2}$] & 89$\pm$02 & \textbf{90$\pm$04} & 68$\pm$06 & 80$\pm$10 & 84$\pm$07 & 87$\pm$02 & 79$\pm$07 & 75$\pm$09 \\
     & ARI [$10^{-2}$] & 74$\pm$06 & \textbf{75$\pm$08} & 45$\pm$14 & 65$\pm$08 & 67$\pm$09 & 70$\pm$05 & 59$\pm$09 & 57$\pm$11 \\
     & NMI [$10^{-2}$] & 76$\pm$03 & \textbf{78$\pm$05} & 56$\pm$16 & 71$\pm$04 & 73$\pm$05 & 72$\pm$04 & 65$\pm$08 & 65$\pm$10 \\
     & Time [$10^{-3}{\mathrm{s}}$] & 35$\pm$02 & 35$\pm$02 & 39$\pm$04 & 30$\pm$02 & 48$\pm$04 & 02$\pm$00 & 01$\pm$00 & \textbf{00$\pm$00} \\
    \midrule
    \multirow{4}{*}{\shortstack[l]{{KDD99}\\$N=494021$\\$\vardim=121$}} & ACC [$10^{-2}$] & 98$\pm$00 & 98$\pm$00 & 98$\pm$00 & 98$\pm$00 & \textbf{98$\pm$00} & 98$\pm$00 & 92$\pm$06 & 91$\pm$07 \\
     & ARI [$10^{-2}$] & 93$\pm$01 & 93$\pm$01 & 94$\pm$00 & 93$\pm$02 & \textbf{94$\pm$01} & 92$\pm$02 & 62$\pm$34 & 59$\pm$36 \\
     & NMI [$10^{-2}$] & 87$\pm$03 & 87$\pm$03 & 88$\pm$01 & 85$\pm$04 & \textbf{88$\pm$02} & 83$\pm$04 & 54$\pm$31 & 51$\pm$32 \\
     & Time [$10^{0}{\mathrm{s}}$] & 04$\pm$00 & 04$\pm$00 & 14$\pm$00 & \textbf{02$\pm$00} & 26$\pm$00 & 56$\pm$16 & 09$\pm$00 & 02$\pm$00 \\
    \midrule
    \multirow{4}{*}{\shortstack[l]{{Leaf}\\$N=340$\\$\vardim=15$}} & ACC [$10^{-2}$] & \textbf{48$\pm$02} & 47$\pm$02 & 36$\pm$02 & 43$\pm$01 & 46$\pm$02 & 40$\pm$02 & 28$\pm$01 & 23$\pm$03 \\
     & ARI [$10^{-2}$] & \textbf{30$\pm$02} & 30$\pm$02 & 20$\pm$02 & 27$\pm$01 & 28$\pm$01 & 25$\pm$02 & 16$\pm$01 & 10$\pm$03 \\
     & NMI [$10^{-2}$] & 64$\pm$01 & \textbf{64$\pm$01} & 56$\pm$02 & 62$\pm$01 & 63$\pm$01 & 58$\pm$01 & 51$\pm$02 & 46$\pm$04 \\
     & Time [$10^{-2}{\mathrm{s}}$] & 07$\pm$00 & 07$\pm$00 & 09$\pm$00 & 03$\pm$00 & 12$\pm$00 & 28$\pm$07 & 01$\pm$00 & \textbf{00$\pm$00} \\
    \midrule
    \multirow{4}{*}{\shortstack[l]{{Libras}\\$N=360$\\$\vardim=90$}} & ACC [$10^{-2}$] & \textbf{47$\pm$01} & \textbf{47$\pm$01} & 38$\pm$03 & 43$\pm$03 & 45$\pm$01 & 29$\pm$02 & 30$\pm$03 & 34$\pm$03 \\
     & ARI [$10^{-2}$] & 31$\pm$02 & \textbf{33$\pm$02} & 20$\pm$03 & 29$\pm$03 & 28$\pm$02 & 16$\pm$03 & 15$\pm$04 & 18$\pm$03 \\
     & NMI [$10^{-2}$] & 57$\pm$02 & \textbf{59$\pm$02} & 46$\pm$03 & 55$\pm$03 & 55$\pm$02 & 38$\pm$04 & 38$\pm$04 & 42$\pm$04 \\
     & Time [$10^{-2}{\mathrm{s}}$] & 05$\pm$00 & 05$\pm$00 & 09$\pm$01 & 03$\pm$00 & 09$\pm$00 & 14$\pm$04 & 00$\pm$00 & \textbf{00$\pm$00} \\
    \midrule
    \multirow{4}{*}{\shortstack[l]{{MNIST}\\$N=60000$\\$\vardim=784$}} & ACC [$10^{-2}$] & 57$\pm$02 & \textbf{59$\pm$01} & 47$\pm$05 & 55$\pm$02 & 43$\pm$04 & 30$\pm$03 & 38$\pm$03 & 35$\pm$04 \\
     & ARI [$10^{-2}$] & 38$\pm$03 & \textbf{42$\pm$02} & 26$\pm$04 & 38$\pm$02 & 20$\pm$04 & 12$\pm$02 & 19$\pm$03 & 14$\pm$04 \\
     & NMI [$10^{-2}$] & 48$\pm$01 & \textbf{53$\pm$02} & 38$\pm$04 & 49$\pm$02 & 32$\pm$03 & 22$\pm$02 & 28$\pm$02 & 26$\pm$04 \\
     & Time [$10^{-1}{\mathrm{s}}$] & 16$\pm$00 & 16$\pm$00 & 42$\pm$02 & 08$\pm$00 & 58$\pm$01 & 45$\pm$02 & 15$\pm$01 & \textbf{05$\pm$01} \\
    \midrule
    \multirow{4}{*}{\shortstack[l]{{Seeds}\\$N=210$\\$\vardim=7$}} & ACC [$10^{-2}$] & \textbf{89$\pm$02} & 87$\pm$03 & 80$\pm$10 & 87$\pm$06 & 83$\pm$05 & 87$\pm$03 & 76$\pm$06 & 69$\pm$08 \\
     & ARI [$10^{-2}$] & \textbf{71$\pm$04} & 68$\pm$06 & 60$\pm$12 & 68$\pm$08 & 60$\pm$08 & 67$\pm$06 & 53$\pm$07 & 48$\pm$11 \\
     & NMI [$10^{-2}$] & \textbf{69$\pm$02} & 67$\pm$03 & 62$\pm$07 & 67$\pm$04 & 63$\pm$05 & 65$\pm$04 & 58$\pm$05 & 54$\pm$07 \\
     & Time [$10^{-3}{\mathrm{s}}$] & 45$\pm$44 & 45$\pm$44 & 52$\pm$05 & 29$\pm$02 & 49$\pm$07 & 02$\pm$00 & 01$\pm$00 & \textbf{00$\pm$00} \\
    \midrule
    \multirow{4}{*}{\shortstack[l]{{Texture}\\$N=5500$\\$\vardim=40$}} & ACC [$10^{-2}$] & \textbf{63$\pm$03} & 63$\pm$02 & 54$\pm$04 & 58$\pm$03 & 62$\pm$03 & 53$\pm$03 & 28$\pm$02 & 37$\pm$06 \\
     & ARI [$10^{-2}$] & 49$\pm$02 & \textbf{49$\pm$02} & 39$\pm$05 & 45$\pm$02 & 47$\pm$03 & 38$\pm$02 & 18$\pm$01 & 24$\pm$06 \\
     & NMI [$10^{-2}$] & 64$\pm$01 & \textbf{65$\pm$01} & 58$\pm$03 & 62$\pm$01 & 63$\pm$02 & 54$\pm$02 & 38$\pm$02 & 46$\pm$05 \\
     & Time [$10^{-2}{\mathrm{s}}$] & 13$\pm$01 & 13$\pm$01 & 21$\pm$01 & 06$\pm$00 & 41$\pm$01 & 72$\pm$23 & 10$\pm$04 & \textbf{02$\pm$00} \\
    \midrule
    \multirow{4}{*}{\shortstack[l]{{Thyroid}\\$N=215$\\$\vardim=5$}} & ACC [$10^{-2}$] & \textbf{87$\pm$05} & 86$\pm$06 & 77$\pm$03 & 84$\pm$04 & 85$\pm$05 & 82$\pm$06 & 73$\pm$05 & 79$\pm$05 \\
     & ARI [$10^{-2}$] & \textbf{59$\pm$16} & 54$\pm$19 & 29$\pm$11 & 50$\pm$13 & 54$\pm$16 & 46$\pm$19 & 20$\pm$19 & 34$\pm$17 \\
     & NMI [$10^{-2}$] & \textbf{57$\pm$13} & 53$\pm$16 & 30$\pm$10 & 49$\pm$11 & 51$\pm$14 & 46$\pm$17 & 23$\pm$20 & 36$\pm$16 \\
     & Time [$10^{-3}{\mathrm{s}}$] & 47$\pm$05 & 47$\pm$05 & 50$\pm$08 & 28$\pm$02 & 48$\pm$04 & 04$\pm$02 & 01$\pm$00 & \textbf{00$\pm$00} \\
    \midrule
    \multirow{4}{*}{\shortstack[l]{{WDBC}\\$N=569$\\$\vardim=30$}} & ACC [$10^{-2}$] & 92$\pm$02 & 91$\pm$02 & 89$\pm$08 & 91$\pm$01 & 86$\pm$09 & \textbf{92$\pm$00} & 87$\pm$07 & 75$\pm$11 \\
     & ARI [$10^{-2}$] & 71$\pm$07 & 69$\pm$08 & 63$\pm$23 & 69$\pm$06 & 55$\pm$23 & \textbf{72$\pm$02} & 58$\pm$20 & 27$\pm$27 \\
     & NMI [$10^{-2}$] & \textbf{61$\pm$05} & 60$\pm$06 & 55$\pm$19 & 60$\pm$04 & 47$\pm$20 & 60$\pm$02 & 48$\pm$18 & 24$\pm$23 \\
     & Time [$10^{-3}{\mathrm{s}}$] & 58$\pm$02 & 58$\pm$02 & 77$\pm$13 & 28$\pm$03 & 58$\pm$08 & 05$\pm$06 & 02$\pm$00 & \textbf{01$\pm$00} \\
    \midrule
    \multirow{4}{*}{\shortstack[l]{{Wine}\\$N=178$\\$\vardim=13$}} & ACC [$10^{-2}$] & \textbf{94$\pm$01} & 92$\pm$04 & 78$\pm$14 & 90$\pm$08 & 83$\pm$09 & 85$\pm$06 & 74$\pm$09 & 66$\pm$13 \\
     & ARI [$10^{-2}$] & \textbf{83$\pm$04} & 79$\pm$10 & 56$\pm$20 & 77$\pm$12 & 62$\pm$14 & 62$\pm$12 & 47$\pm$13 & 36$\pm$18 \\
     & NMI [$10^{-2}$] & \textbf{81$\pm$03} & 79$\pm$07 & 61$\pm$15 & 77$\pm$08 & 66$\pm$09 & 66$\pm$09 & 53$\pm$12 & 42$\pm$17 \\
     & Time [$10^{-3}{\mathrm{s}}$] & 36$\pm$01 & 36$\pm$01 & 45$\pm$05 & 31$\pm$02 & 46$\pm$02 & 07$\pm$03 & 01$\pm$00 & \textbf{00$\pm$00} \\
    \bottomrule
  \end{tabular}
\end{table}

\par Tables~\ref{tab:clustering_synth}--\ref{tab:clustering_real} report clustering performance and runtime. We consider eFDPC-$\mathcal{P}$, which clusters using the selected peaks only, and eFDPC-$\mathcal{G}$, which additionally propagates peak labels to the remaining micro-clusters at the server. Best results are highlighted in bold.
Overall, both eFDPC variants are consistently among the best-performing methods across datasets, with particularly strong gains on datasets with overlap, non-spherical structure, or varying density. In contrast, faster baselines often trade accuracy for speed, while some methods degrade substantially when the number of clusters/classes is large or when cluster shapes are not axis-aligned. On the synthetic datasets (Table~\ref{tab:clustering_synth}), eFDPC achieves the best performance on most benchmarks, especially on datasets with pronounced density variations or overlap. In particular, on Overlap, S3--S4, D31, and R15 density-peak selection results in notably higher ARI and NMI, because operating on incrementally learned micro-clusters reduces the chance of missing true density peaks. The selection then primarily depends on whether candidate peaks have high enough local density and sufficient separation from higher-density regions. On real-world datasets (Table~\ref{tab:clustering_real}), eFDPC achieves the best overall performance on several datasets, with the highest ACC and typically the highest ARI and NMI. On KDD99, which has large $N$ and high $\vardim$, {$k$}-means-based baselines ($k$-FED and FKDC) sometimes fail to recover meaningful cluster structure, whereas eFDPC maintains high ACC and achieves higher ARI and NMI. The fuzzy $C$-means-based FedFCM achieves very strong results, but with an order-of-magnitude higher runtime. On high-dimensional {MNIST} ($\vardim=784$), eFDPC-$\mathcal{G}$ substantially improves over most baselines in ACC and NMI, while $k$-FED achieves a comparable ARI.

\par In terms of runtime, eFDPC is slower than the fastest centroid-based baselines (e.g., $k$-FED and LLF-CM/LLF-FCM), but it remains substantially faster than heavier approaches such as FedFCM on large-scale datasets, while providing stronger ARI and NMI. This reflects the added cost of forming and merging micro-clusters and performing peak selection, which improves robustness at a moderate computational overhead. However, the reported runtimes measure local computation only and do not include 
the communication cost for any method, which would add overhead in a real deployment.

\subsection{Federated learning sensitivity}
\par We evaluate the sensitivity of eFDPC under changing federated conditions and learning settings. We vary the non-IID level of client data distributions with a Dirichlet $\alpha$ sweep, we vary the amount of data per client with a sweep over the number of clients, {and we vary the mini-batch size used by the client-side evolving learning}. All other settings are the same as in the benchmark-dataset experiments.

\par First, we performed a client-count sweep to test how clustering changes when the data are distributed among more clients. As $K$ increases, each client observes fewer samples and fewer classes, so local models become less representative, and server-side aggregation must compute peaks from more partial and more heterogeneous client contributions. We keep the experiment setup identical to the benchmark-dataset experiments and vary only the number of clients, $K\in\{5,10,30\}$.

\par The results of the client-count sweep are presented in Table~\ref{tab:mnist_clientsweep}. The metrics stay stable or improve as $K$ increases. The end-to-end runtime drops for most methods when the data are split across more clients because each client processes a smaller local partition. When the data are partitioned among more clients, each client processes fewer samples and maintains fewer micro-clusters than in a centralized setting. Local distance computations scale as $\mathcal{O}(C^{{(k)}} N^{{(k)}})$ and pairwise cluster operations scale as $\mathcal{O}\big({(C^{(k)})^2 N^{(k)}}\big)$. Under an approximately balanced split, $N^{{(k)}}\approx N/K$ and $C^{{(k)}}\approx C/K$, so the per-client cost decreases to $\mathcal{O}(CN/K^2)$ and $\mathcal{O}(C^2/K^2)$, and the total work across clients decreases roughly as $\mathcal{O}(CN/K)$ and $\mathcal{O}(C^2/K)$. The increase in metrics for larger $K$ is expected because every compared method sends a fixed number of clusters per client, so a larger $K$ means that the server receives more clusters in total. This makes aggregation easier and can improve ACC, ARI, and NMI across the board.

\begin{table}[!t]
  \centering
  \footnotesize
  \caption{Client-count $K$ sweep for Dirichlet $\alpha = 0.9${.}}\label{tab:mnist_clientsweep}
  \setlength{\tabcolsep}{2pt}
  \begin{tabular}{l|l|llllllll}
    \toprule
    Dataset & Metric & eFDPC-$\mathcal{P}$ & eFDPC-$\mathcal{G}$ & FKDC & $k$-FED & MUFC & FedFCM & LLF-FCM & LLF-CM \\
    \midrule
   \multirow{4}{*}{\shortstack[l]{{MNIST}\\$K=5$}} & ACC [$10^{-2}$] & 52$\pm$02 & \textbf{59$\pm$04} & 55$\pm$02 & 55$\pm$03 & 45$\pm$03 & 31$\pm$04 & 38$\pm$03 & 35$\pm$04 \\
     & ARI [$10^{-2}$] & 28$\pm$03 & \textbf{39$\pm$05} & 36$\pm$02 & 38$\pm$02 & 24$\pm$04 & 13$\pm$04 & 19$\pm$03 & 14$\pm$04 \\
     & NMI [$10^{-2}$] & 42$\pm$01 & \textbf{55$\pm$03} & 46$\pm$02 & 50$\pm$02 & 35$\pm$03 & 23$\pm$05 & 28$\pm$02 & 26$\pm$04 \\
     & Time [$10^{-1}{\mathrm{s}}$] & 16$\pm$00 & 16$\pm$00 & 43$\pm$01 & 07$\pm$00 & 55$\pm$00 & 50$\pm$27 & 15$\pm$00 & \textbf{05$\pm$01} \\
     \midrule
    \multirow{4}{*}{\shortstack[l]{{MNIST}\\$K=10$}} & ACC [$10^{-2}$] & 51$\pm$03 & \textbf{60$\pm$04} & 54$\pm$02 & 55$\pm$05 & 47$\pm$02 & 33$\pm$03 & 36$\pm$03 & 33$\pm$06 \\
     & ARI [$10^{-2}$] & 29$\pm$03 & \textbf{43$\pm$06} & 32$\pm$02 & 39$\pm$04 & 27$\pm$03 & 13$\pm$03 & 17$\pm$02 & 11$\pm$04 \\
     & NMI [$10^{-2}$] & 42$\pm$02 & \textbf{58$\pm$03} & 42$\pm$02 & 49$\pm$04 & 38$\pm$02 & 24$\pm$04 & 26$\pm$02 & 24$\pm$06 \\
     & Time [$10^{0}{\mathrm{s}}$] & 10$\pm$08 & 10$\pm$08 & 04$\pm$01 & 00$\pm$00 & 06$\pm$00 & 04$\pm$00 & 01$\pm$00 & \textbf{00$\pm$00} \\
    \midrule
    \multirow{4}{*}{\shortstack[l]{{MNIST}\\$K=30$}} & ACC [$10^{-2}$] & 55$\pm$04 & \textbf{60$\pm$04} & 48$\pm$02 & 57$\pm$03 & 50$\pm$02 & 34$\pm$03 & 38$\pm$03 & 35$\pm$04 \\
     & ARI [$10^{-2}$] & 33$\pm$04 & \textbf{41$\pm$05} & 26$\pm$02 & 40$\pm$02 & 30$\pm$02 & 15$\pm$02 & 18$\pm$03 & 14$\pm$03 \\
     & NMI [$10^{-2}$] & 45$\pm$03 & \textbf{55$\pm$03} & 36$\pm$02 & 51$\pm$01 & 41$\pm$02 & 26$\pm$02 & 27$\pm$02 & 26$\pm$03 \\
     & Time [$10^{0}{\mathrm{s}}$] & 01$\pm$00 & 01$\pm$00 & 08$\pm$00 & 00$\pm$00 & 05$\pm$00 & 04$\pm$00 & 01$\pm$00 & \textbf{00$\pm$00} \\
    \bottomrule
  \end{tabular}
\end{table}

\par {Second, we performed a mini-batch size sweep to test how the batch size $B$ affects model quality. The batch size controls how many samples are processed in one update. Small batches update the model more frequently, but they are slower because data-loading overheads occur more often. This also generally produces fewer clusters. In contrast, large batches reduce the number of updates, but they can produce too many clusters during the early mini-batches. We tested $B\in\{32,64,128,256,512,1024\}$ and the number of clients $K\in\{3,10,30\}$ at $\alpha=0.9$ for both the peak-only variant eFDPC-$\mathcal{P}$ and the grouped variant eFDPC-$\mathcal{G}$.}

\par
\begin{figure}[!t]
    \centering
    \begin{minipage}[t]{0.45\linewidth}\centering
        \includegraphics[width=\linewidth]{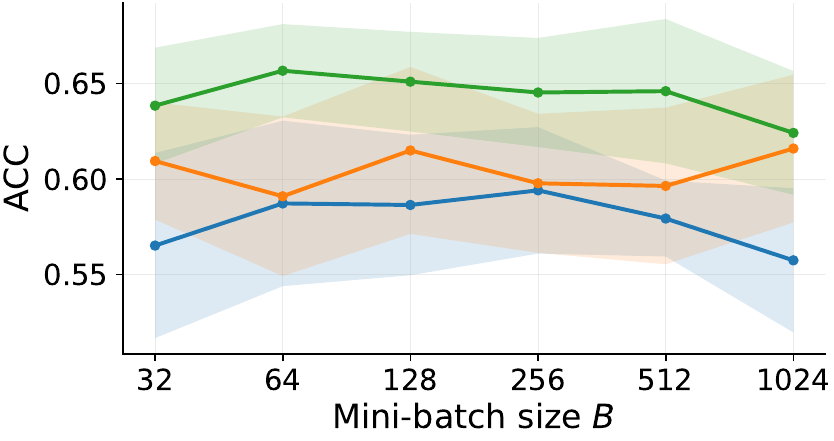}
        {\footnotesize{(a)~Accuracy (ACC)}}
    \end{minipage}%
    \begin{minipage}[t]{0.45\linewidth}\centering
        \includegraphics[width=\linewidth]{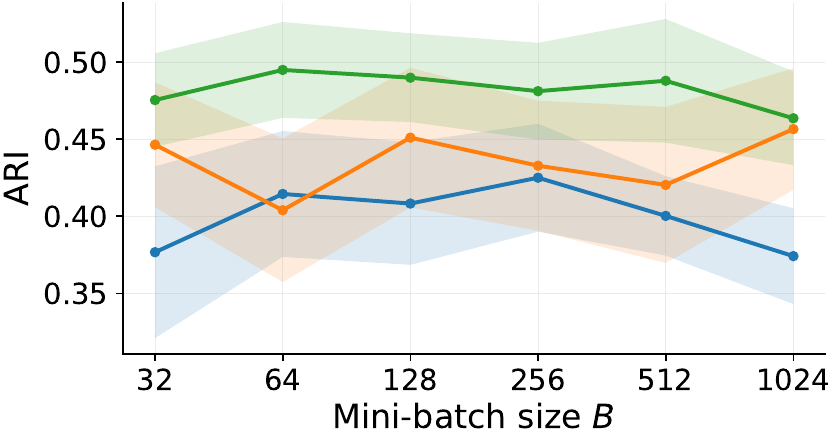}
        {\footnotesize{(b)~Adjusted Rand Index (ARI)}}
    \end{minipage}

    \begin{minipage}[t]{0.45\linewidth}\centering
        \includegraphics[width=\linewidth]{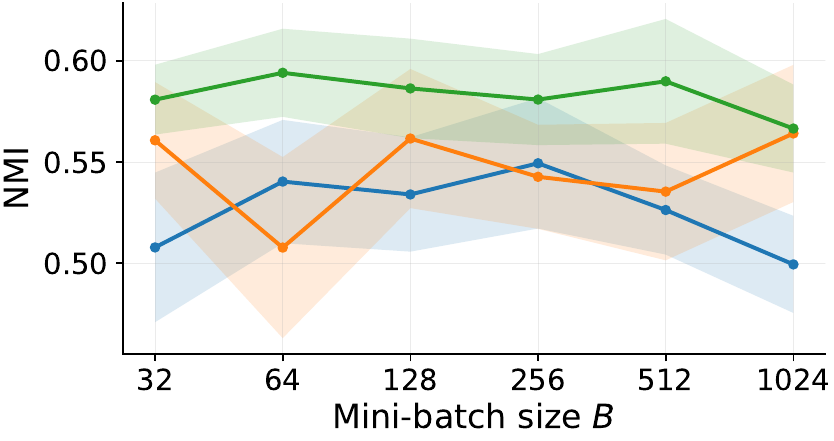}
        {\footnotesize{(c)~Normalized Mutual Information (NMI)}}
    \end{minipage}%
    \begin{minipage}[t]{0.45\linewidth}\centering
        \includegraphics[width=\linewidth]{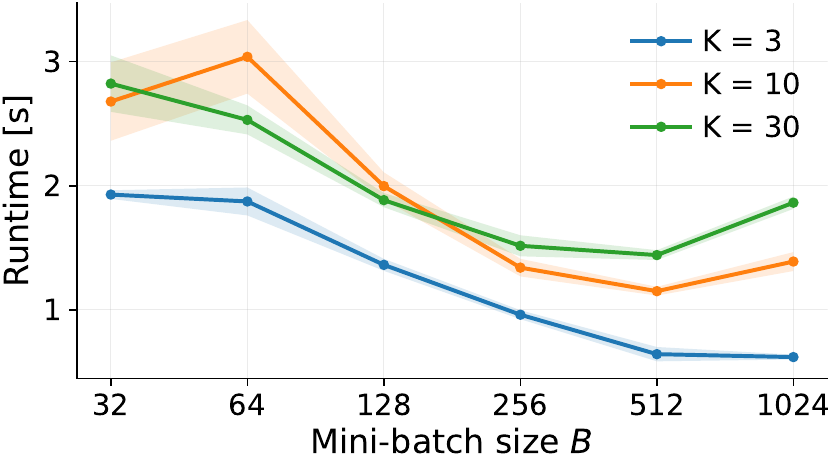}
        {\footnotesize{(d)~Wall-clock runtime}}
    \end{minipage}

    \caption{{Mini-batch size sensitivity of eFDPC-$\mathcal{G}$ on MNIST for
    $K=3,10,30$ clients (Dirichlet $\alpha=0.9$). ACC, ARI, and NMI are stable across the
    mini-batch size $B$, with a mild optimum at moderate sizes, and improve with the number
    of clients; runtime falls as $B$ grows and grows with $K$. Bands show $\pm1$ standard
    deviation over the repetitions.}}
    \label{fig_batchsweep}
\end{figure}

 {The results of the sweep experiment are shown in Fig.~\ref{fig_batchsweep}. 
Clustering quality is stable across a broad range of batch sizes, with an optimum around $B=64$ to $128$, and a small degradation at the largest batch ($B=1024$), \emph{i.e.}, the default $B\approx\sqrt{N}$ therefore lies within this stable region. Local runtime falls as $B$ grows, because fewer mini-batches are processed, so larger batches reduce computation at little cost in accuracy. Increasing the number of clients improves accuracy, with eFDPC-$\mathcal{G}$ rising from about $0.59$ at $K=3$ to $0.66$ at $K=30$, consistent with the client-count sweep in the previous experiment. In contrast to the per-client cost, the eFDPC runtime grows modestly with $K$, because the server-side reconstruction and global density peak clustering operate on the larger pooled set of $KP$ peaks at cost $\mathcal{O}((KP)^2 d)$ (Table~\ref{tab_cost}). Across all settings eFDPC-$\mathcal{G}$ outperforms eFDPC-$\mathcal{P}$, confirming the benefit of the server-side label propagation.}

\par {Third}, we performed a non-IID Dirichlet $\alpha$ sweep to test how clustering changes as the client data split becomes more or less non-IID. Smaller $\alpha$ gives a more skewed split, where each client has samples from fewer classes, while larger $\alpha$ produces a more balanced split across clients. We keep the experiment setup identical to the benchmark-dataset experiments and vary only the Dirichlet concentration parameter, using $\alpha\in\{0.1,0.3,1,3.0\}$ with $K=10$ clients.

\par The results of the $\alpha$ sweep experiment are shown in Table~\ref{tab:mnist_alphasweep}. As $\alpha$ increases, eFDPC-$\mathcal{G}$ improves consistently in ACC, ARI, and NMI, which matches the expectation that a less skewed client split makes server-side peak reconstruction and grouping easier. At lower $\alpha$, clients agree less on which micro-clusters form global peaks, so more non-peak micro-clusters are shared and the peaks become less pronounced, which can lead to less stable selection. Across all tested $\alpha$ values, eFDPC-$\mathcal{G}$ achieves the strongest clustering quality among the compared methods. Runtime is broadly comparable across $\alpha$ for all methods. A likely reason for higher runtime at larger $\alpha$ is that micro-clusters from different clients become closer, so the server keeps more neighbor edges and evaluates more candidate pairs during merging and grouping even when the number of micro-clusters $C$ is similar.

\begin{figure}[!t]
    \centering
    \begin{minipage}[t]{0.245\linewidth}\centering
        \includegraphics[width=\linewidth]{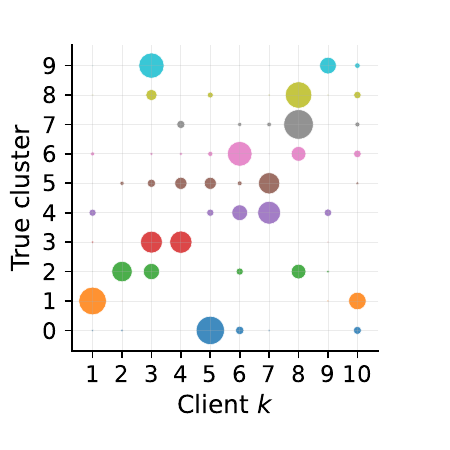}\\[-5pt]
        {\footnotesize (a)~$\alpha=0.1$}
    \end{minipage}\hfill
    \begin{minipage}[t]{0.245\linewidth}\centering
        \includegraphics[width=\linewidth]{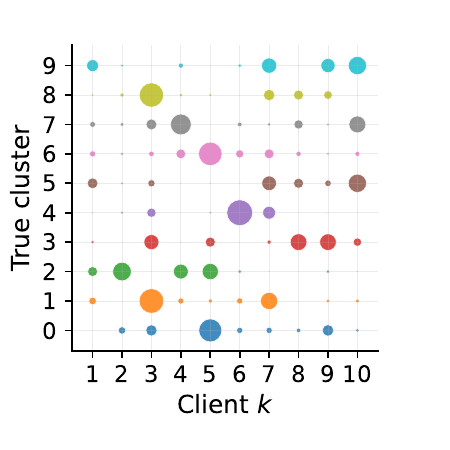}\\[-5pt]
        {\footnotesize (b)~$\alpha=0.3$}
    \end{minipage}\hfill
    \begin{minipage}[t]{0.245\linewidth}\centering
        \includegraphics[width=\linewidth]{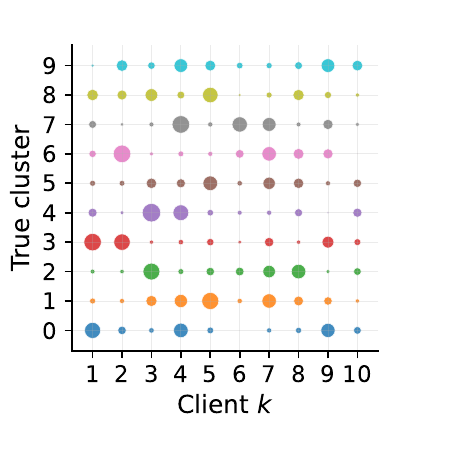}\\[-5pt]
        {\footnotesize (c)~$\alpha=1$}
    \end{minipage}\hfill
    \begin{minipage}[t]{0.245\linewidth}\centering
        \includegraphics[width=\linewidth]{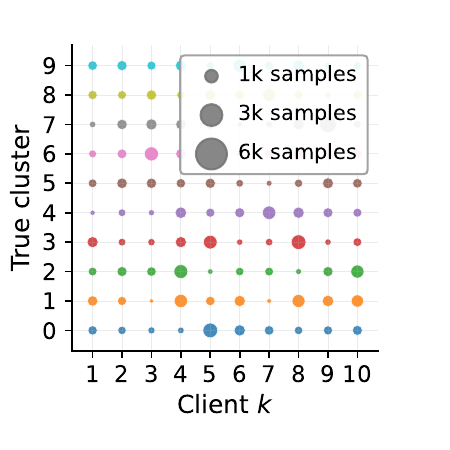}\\[-5pt]
        {\footnotesize (d)~$\alpha=3.0$}
    \end{minipage}

    \caption{{Non-IID label partition of MNIST across $K=10$ clients with the
    Dirichlet split used in the experiments. Small $\alpha$ results in each true
    cluster being split between a few clients (a), while large $\alpha$ approaches an IID split where every
    client holds every cluster in similar proportions (d).}}
    \label{fig:noniid_alpha}
\end{figure}

\begin{table}[!t]
  \centering
  \footnotesize
  \caption{Non-IID Dirichlet $\alpha$ sweep for $K=10${.}}\label{tab:mnist_alphasweep}
  \setlength{\tabcolsep}{2pt}
  \begin{tabular}{l|l|llllllll}
    \toprule
    Dataset & Metric & eFDPC-$\mathcal{P}$ & eFDPC-$\mathcal{G}$ & FKDC & $k$-FED & MUFC & FedFCM & LLF-FCM & LLF-CM \\
    \midrule
    \multirow{4}{*}{\shortstack[l]{{MNIST}\\$\alpha=0.1$}} & ACC [$10^{-2}$] & 48$\pm$05 & \textbf{54$\pm$06} & 53$\pm$04 & 48$\pm$06 & 48$\pm$03 & 30$\pm$06 & 38$\pm$03 & 35$\pm$04 \\
     & ARI [$10^{-2}$] & 29$\pm$04 & \textbf{37$\pm$07} & 33$\pm$04 & 31$\pm$06 & 26$\pm$04 & 11$\pm$05 & 19$\pm$03 & 14$\pm$04 \\
     & NMI [$10^{-2}$] & 38$\pm$04 & \textbf{52$\pm$05} & 43$\pm$04 & 44$\pm$05 & 38$\pm$03 & 20$\pm$06 & 28$\pm$02 & 26$\pm$04 \\
     & Time [$10^{0}{\mathrm{s}}$] & 03$\pm$03 & 03$\pm$03 & 04$\pm$00 & 00$\pm$00 & 06$\pm$00 & 12$\pm$14 & 01$\pm$00 & \textbf{00$\pm$00} \\
    \midrule
    \multirow{4}{*}{\shortstack[l]{{MNIST}\\$\alpha=0.3$}} & ACC [$10^{-2}$] & 49$\pm$04 & \textbf{57$\pm$05} & 54$\pm$03 & 52$\pm$04 & 48$\pm$02 & 33$\pm$03 & 38$\pm$03 & 35$\pm$04 \\
     & ARI [$10^{-2}$] & 28$\pm$04 & \textbf{41$\pm$06} & 33$\pm$03 & 35$\pm$03 & 27$\pm$02 & 14$\pm$03 & 19$\pm$03 & 14$\pm$04 \\
     & NMI [$10^{-2}$] & 39$\pm$03 & \textbf{56$\pm$04} & 43$\pm$03 & 47$\pm$03 & 38$\pm$02 & 24$\pm$04 & 28$\pm$02 & 26$\pm$04 \\
     & Time [$10^{0}{\mathrm{s}}$] & 02$\pm$00 & 02$\pm$00 & 05$\pm$00 & 00$\pm$00 & 06$\pm$00 & 05$\pm$04 & 01$\pm$00 & \textbf{00$\pm$00} \\
    \midrule
    \multirow{4}{*}{\shortstack[l]{{MNIST}\\$\alpha=1.0$}} & ACC [$10^{-2}$] & 50$\pm$04 & \textbf{61$\pm$05} & 55$\pm$03 & 57$\pm$03 & 46$\pm$03 & 32$\pm$03 & 38$\pm$03 & 35$\pm$04 \\
     & ARI [$10^{-2}$] & 28$\pm$05 & \textbf{44$\pm$06} & 34$\pm$03 & 39$\pm$02 & 27$\pm$03 & 13$\pm$03 & 19$\pm$03 & 14$\pm$04 \\
     & NMI [$10^{-2}$] & 41$\pm$03 & \textbf{60$\pm$03} & 44$\pm$02 & 50$\pm$02 & 37$\pm$02 & 24$\pm$03 & 28$\pm$02 & 26$\pm$04 \\
     & Time [$10^{0}{\mathrm{s}}$] & 03$\pm$00 & 03$\pm$00 & 05$\pm$00 & 00$\pm$00 & 07$\pm$00 & 03$\pm$00 & 02$\pm$00 & \textbf{00$\pm$00} \\
    \midrule
    \multirow{4}{*}{\shortstack[l]{{MNIST}\\$\alpha=3.0$}} & ACC [$10^{-2}$] & 51$\pm$03 & \textbf{63$\pm$04} & 53$\pm$02 & 56$\pm$02 & 47$\pm$02 & 30$\pm$03 & 38$\pm$03 & 35$\pm$04 \\
     & ARI [$10^{-2}$] & 29$\pm$04 & \textbf{48$\pm$05} & 33$\pm$02 & 39$\pm$02 & 26$\pm$02 & 12$\pm$03 & 19$\pm$03 & 14$\pm$04 \\
     & NMI [$10^{-2}$] & 41$\pm$03 & \textbf{61$\pm$03} & 42$\pm$01 & 50$\pm$02 & 37$\pm$02 & 22$\pm$03 & 28$\pm$02 & 27$\pm$04 \\
     & Time [$10^{0}{\mathrm{s}}$] & 10$\pm$07 & 10$\pm$07 & 05$\pm$00 & 00$\pm$00 & 07$\pm$00 & 03$\pm$00 & 02$\pm$00 & \textbf{00$\pm$00} \\
    \bottomrule
  \end{tabular}
\end{table}

\section{Discussion}\label{sec:discussion}
\par The experimental results indicate that eFDPC-$\mathcal{P}$ is consistently among the best-performing methods across datasets, with particularly strong gains on datasets with overlap and varying density, while eFDPC-$\mathcal{G}$ performs especially well on non-spherical structure. In this section, we discuss why the proposed micro-cluster representation and peak-selection strategy are effective under federated constraints, and we highlight limitations and potential extensions.

\par The main idea of eFDPC is to form clusters from axis-parallel (diagonal-covariance) micro-clusters, which have substantially lower computational cost, and to retain much of the descriptive power of full-covariance Gaussian models used in evolving fuzzy systems~\cite{Ozbot_Ozawa_skrjanc_2025, Lughofer_Cernuda_Kindermann_Pratama_2015, Skrjanc_Iglesias_Sanchis_Leite_Lughofer_Gomide_2019} through density-peak selection and peak-based label propagation. Full-covariance clusters have an advantage over axis-parallel (diagonal) clusters because they model correlations between features, which can improve accuracy and/or reduce the number of required clusters compared to spherical models~\cite{Lughofer_Cernuda_Kindermann_Pratama_2015}. However, a full covariance matrix has $\vardim(\vardim+1)/2$ free parameters, and reliable estimation requires at least that number of samples per cluster, whereas high-dimensional data are often sparse. Consequently, for high-dimensional data, a diagonal covariance with $\vardim$ parameters, or even a spherical model, often performs better while requiring far less computation~\cite{Bouveyron_Girard_Schmid_2007,Ozbot_Ozawa_skrjanc_2025}. Further, the time complexity to compute the inverse of a covariance matrix used in distance and overlap computations is $\mathcal{O}(\vardim^3)$, while for diagonal covariances it is $\mathcal{O}(\vardim)$, implying that multivariate methods are typically much slower than methods that rely on centers and variances only.

\par A key component is overlap-aware merging during micro-cluster evolution. Existing dissimilarity measures used in evolving clustering suffer from several limitations. First, measures based on covariance determinants or explicit volume computation are computationally expensive, especially with multivariate Gaussian covariances, since determinant and inverse operations are costly and pairwise comparisons lead to high overall cost~\cite{skrjanc_2020}. Second, many measures do not include sample counts, so clusters with different support are treated equally~\cite{Ozbot_Ozawa_Skrjanc_2024}. Third, measures that rely on the geometric mean of covariance determinants, such as the Bhattacharyya distance~\cite{Ozbot_Skrjanc_eais_2024, Lughofer_Cernuda_Kindermann_Pratama_2015}, can fail to detect overlap between small and large clusters because they measure absolute overlap, which can be small even when clusters share the same center. In eFDPC, the overlap-aware merging criterion is designed to account for both dispersion and relative support, which reduces premature merging and improves robustness when clusters are skewed, overlapping, or have very dissimilar densities. 

\par The eFDPC clients do not transmit absolute micro-cluster {supports}, as these may leak additional information about local data. Interestingly, avoiding absolute counts can also improve the quality of density peak selection. The resulting peak set is less dominated by large local densities, \emph{i.e.}, dense regions do not overwhelm the final model, and lower-density peaks can still contribute meaningfully, \emph{e.g.}, in the Unbalance and KDD99 datasets. Further, using bounded {supports $\tilde n_c^{(k)}\!\in[0,1]$} together with equal treatment of transmitted centers acts as a soft voting mechanism across clients, so each client can have comparable influence on the aggregated model even under non-IID partitions. This may reduce the impact of strongly biased local datasets, although defenses against adversarial clients require further investigation. {The number of peaks $P$ is provided in our experiments for a fair comparison with the baselines. $P$ could be selected automatically, for example as an elbow on the sorted decision scores $\gamma_i = \rho_i\,\delta_i$, but no unsupervised criterion recovers the true number of clusters reliably.}

\par The proposed framework also has limitations. First, peak selection and propagation depend on the stability of micro-cluster centers. When the decision graph is flat and no clear peaks exist, grouping can become sensitive to local perturbations and neighborhood construction. Second, the overall quality depends on hyperparameters such as the merging aggressiveness and neighborhood size used to localize density or propagation, which must be selected consistently across datasets. Third, while micro-cluster peaks reduce communication compared to transmitting samples, they still introduce communication overhead relative to pure centroid-only baselines, and the shuffler-based anonymization layer adds system complexity and communication overhead. Finally, propagation can blur boundaries when micro-clusters are mixed. For applications where strict boundary separation is critical, peak-only grouping (eFDPC-$\mathcal{P}$) may be preferable. eFDPC-$\mathcal{G}$ is more suitable when the data are not well described by centroid-based clusters and instead lie on a lower-dimensional manifold. {Because the server reconstructs global peaks from all clients, a peak missed by one client is recovered from another (Fig. 3). The cluster is missed only when a true cluster is under-represented for all clients and no client transmits a peak for it.}

\par The peaks computed with DPC are derived from micro-cluster centers, so peak selection can be sensitive to perturbations in micro-cluster assignments during cluster evolution. If a label propagation stage is used, it can also reduce clustering accuracy, since boundary micro-clusters may contain samples from multiple clusters and can shift the assignment balance. Rather than assigning each micro-cluster to its nearest higher-density neighbor as in classical DPC, we propagate labels simultaneously from all selected peaks over $E_\kappa$. This results in a partition that follows the local connectivity of $E_\kappa$ and typically produces smoother decision boundaries. In our experiments, nearest-higher propagation was often unstable for micro-clusters, as small shifts in center positions can change nearest-higher relations and produce uneven groups, especially on flatter datasets with no clear density peaks. In contrast, multi-source Dijkstra propagation from peaks was more stable because it depends on local paths in $E_\kappa$ rather than a global density ordering.

\par Privacy could be further increased by applying Differential Privacy (DP). However, applying DP directly to the transmitted centers is not viable in our setting because, for typical clustering datasets and $\varepsilon=1$, the required noise is on the order of the data spread (or larger). As a result, global peak selection becomes unstable and clustering utility degrades substantially. DP is therefore most viable when only securely aggregated information is revealed to the server, \emph{e.g.}, when the server proposes candidates, clients vote, and the votes are securely aggregated. Multi-party computation could also be used to hide the global-peak computation from all parties. However, such secure computation is typically several orders of magnitude slower than the proposed eFDPC pipeline~\cite{Keller_2020}, which makes it difficult to apply at the scale of the considered datasets.

\section{Conclusion}\label{sec:conclusion}
\par This work introduced eFDPC, a one-shot federated clustering framework that combines evolving fuzzy clustering and density peak clustering in a federated setting. Each client evolves micro-clusters from private mini-batches through incremental clustering, cluster addition, and merging, and then transmits only density-peak-selected local representatives through a shuffler-based anonymization layer to enable unlinkable global peak selection at the server. Experiments on synthetic and real-world benchmarks showed that eFDPC consistently ranks among the best-performing approaches under non-IID partitions. It achieves strong ACC and typically improves ARI/NMI on challenging structures such as overlap, varying density, and high dimensionality, while maintaining runtime comparable to {$k$}-means methods. A key limitation of eFDPC is that peak selection and propagation can be sensitive when no clear density peaks exist, and performance depends on quantization and merging hyperparameters. Future work may include extending eFDPC to semi-supervised federated clustering and evolving environments, as well as studying robustness to adversarial clients, privacy mechanisms at the aggregation stage, and communication-aware privacy budgeting under multi-round updates.

%%%

\begin{appendices}
\section{Derivation of the Overlap score}\label{app:kappa_derivation}
\par Consider two multivariate Gaussian clusters $\mathcal{C}_p$ and $\mathcal{C}_q$ with means $\tensor{\mu}_p,\tensor{\mu}_q \in \mathbb{R}^{\vardim}$, covariance matrices $\tensor{\Sigma}_p,\tensor{\Sigma}_q \in \mathbb{R}^{\vardim\times \vardim}$, and {supports} $n_p,n_q$. The goal is to determine if ${p,q}$ should be merged.
\par Based on the multivariate Gaussian merging formulation \cite{Ozbot_Ozawa_Skrjanc_2024}, the covariance matrix of the merged cluster can be split into the average covariance and spread:
\begin{align}
\tensor{\Sigma} &= \tensor{\Sigma}^{\circ} + \tensor{\Sigma}^{-}, \label{eq:cov_decomp}\\
\tensor{\Sigma}^{\circ} &= \frac{n_p \tensor{\Sigma}_p + n_q \tensor{\Sigma}_q}{n_p + n_q}, \label{eq:cov_pooled}\\
\tensor{\Sigma}^{-} &= \frac{n_p n_q}{(n_p+n_q)^2} (\tensor{\mu}_p-\tensor{\mu}_q)(\tensor{\mu}_p-\tensor{\mu}_q)^\top . \label{eq:cov_spread}
\end{align}
Here, $\tensor{\Sigma}^{\circ} \in \mathbb{R}^{\vardim\times \vardim}$ denotes the pooled covariance without spread, while $\tensor{\Sigma}^{-} \in \mathbb{R}^{\vardim\times \vardim}$ is a rank-one positive semidefinite matrix capturing the dispersion induced by the separation between the cluster centers. The hyper-volume of a $\vardim$-dimensional Gaussian cluster with covariance $\tensor{\Sigma} \in \mathbb{R}^{\vardim\times \vardim}$ is proportional to \cite{skrjanc_2020}:
\begin{equation}
V(\tensor{\Sigma}) = \frac{\pi^{\vardim/2}}{\Gamma(\vardim/2 + 1)} \sqrt{\det(\tensor{\Sigma})} . \label{eq:hypervolume}
\end{equation}
and relative volume changes depend only on determinant ratios.
\par We propose to measure dissimilarity based on the relative covariance volume contribution of the center-dispersion term $\tensor{\Sigma}^{-}$. However, this cannot be measured directly because its hyper-volume is 0. Accordingly, we consider the ratio between the hyper-volume of the merged covariance $\tensor{\Sigma}$ and that with the pooled covariance $\tensor{\Sigma}^{\circ}$:
\begin{equation}
\frac{V}{V^{\circ}} = \sqrt{\frac{\det(\tensor{\Sigma})}{\det(\tensor{\Sigma}^{\circ})}} = \sqrt{\det\!\left((\tensor{\Sigma}^{\circ})^{-1}\tensor{\Sigma}\right)} . \label{eq:vol_ratio}
\end{equation}
{We define the overlap score as the squared relative volume ratio:
\begin{equation}
\kappa_{pq} = (V/V^\circ)^2 = \det\!\big((\tensor{\Sigma}^{\circ})^{-1}\tensor{\Sigma}\big) .
\label{eq:overlap_def}
\end{equation}}
Substituting the covariance decomposition in Eq.~\eqref{eq:cov_decomp}:
\begin{equation}
(\tensor{\Sigma}^{\circ})^{-1}\tensor{\Sigma} = \tensor{I} + \frac{n_p n_q}{(n_p+n_q)^2} (\tensor{\Sigma}^{\circ})^{-1} (\tensor{\mu}_p-\tensor{\mu}_q)(\tensor{\mu}_p-\tensor{\mu}_q)^\top . \label{eq:cov_ratio_expanded}
\end{equation}
Using the matrix determinant lemma~\cite{matrix_cookbook}, $\det(\tensor{I}+\tensor{u}\tensor{v}^{\top}) = 1+\tensor{v}^{\top}\tensor{u}$, the determinant of Eq.~\eqref{eq:cov_ratio_expanded} simplifies to:
\begin{equation}
\kappa_{pq} = 1 + \frac{n_p n_q}{(n_p+n_q)^2} (\tensor{\mu}_p-\tensor{\mu}_q)^\top (\tensor{\Sigma}^{\circ})^{-1} (\tensor{\mu}_p-\tensor{\mu}_q). \label{eq:dissimilarity}
\end{equation}
The resulting overlap {score} $\kappa_{pq}$ is symmetric, dimensionless, scales with the cluster spreads, and depends on the cluster means $\tensor{\mu}_p,\tensor{\mu}_q$, covariance matrices $\tensor{\Sigma}_p,\tensor{\Sigma}_q$, and {supports} $n_p,n_q$.

\end{appendices}

\backmatter

\bmhead{Statements and Declarations}

\bmhead{Competing Interests}
The authors declare that they have no competing interests.

\bmhead{Funding}
The authors acknowledge the financial support from the Slovenian Research and Innovation Agency (Ph.D. grant for Miha Ožbot, funding number P2-0219).

\bmhead{Data Availability}
The datasets analyzed during the current study are publicly available from the sources cited in the manuscript.

\bmhead{Code Availability}
The eFDPC PyTorch implementation and the experimental scripts are available at \href{https://github.com/mihaozbot/eFDPC}{github.com/mihaozbot/eFDPC}.

\bmhead{Ethics Approval}
Not applicable.

\bmhead{Consent to Participate}
Not applicable.

\bmhead{Consent for Publication}
Not applicable.

\bibliography{sn-bibliography}% common bib file

\end{document}